\documentclass[runningheads]{llncs}
\usepackage{graphicx}

\usepackage{tikz}
\usepackage{comment}
\usepackage{amsmath,amssymb} 
\usepackage{color}

\usepackage[accsupp]{axessibility}  

\usepackage{orcidlink}

\usepackage{hyperref}       
\hypersetup{colorlinks=true, urlcolor=magenta, citecolor=green}
\usepackage{wrapfig}
\usepackage{url}            
\usepackage{booktabs}       
\usepackage{amsfonts}       
\usepackage{nicefrac}       
\usepackage{microtype}      
\usepackage{graphicx}
\usepackage{amsmath}
\usepackage{pifont}
\usepackage[font=footnotesize,labelfont=bf]{caption}
\usepackage{subcaption}
\usepackage{cuted}
\usepackage{float}
\usepackage{sidecap}

\usepackage{xcolor}
\definecolor{darkred}{rgb}{0.66, 0.13, 0.24}
\definecolor{darkgreen}{rgb}{0.24, 0.13, 0.66}

\usepackage[capitalise,noabbrev]{cleveref}

\usepackage{soul}

\usepackage{xspace}

\makeatletter
\DeclareRobustCommand\onedot{\futurelet\@let@token\@onedot}
\def\@onedot{\ifx\@let@token.\else.\null\fi\xspace}

\newcommand{\dataset}{VTC\xspace}

\def\eg{\emph{e.g}\onedot} 
\def\ie{\emph{i.e}\onedot}

\makeatother

\usepackage{amsmath,amsfonts,bm}

\def\eqref#1{equation~\ref{#1}}

\def\1{\bm{1}}

\newcommand{\test}{\mathcal{D_{\mathrm{test}}}}

\DeclareMathAlphabet{\mathsfit}{\encodingdefault}{\sfdefault}{m}{sl}
\SetMathAlphabet{\mathsfit}{bold}{\encodingdefault}{\sfdefault}{bx}{n}

\newcommand{\midsepchange}{\aboverulesep = 0.0mm \belowrulesep = 0.0mm}
\midsepchange

\begin{document}
\pagestyle{headings}
\mainmatter
\def\ECCVSubNumber{833}  
\title{\dataset: Improving Video-Text Retrieval\\with User Comments} 

\titlerunning{Context-Adapted Video-Text}
%
\author{Laura Hanu\inst{1}\orcidlink{0000-0002-3423-9373} \and
James Thewlis\inst{1}\orcidlink{0000-0001-8410-2570} \and
Yuki M. Asano\inst{2}\orcidlink{0000-0002-8533-4020} \and 
Christian Rupprecht\inst{3}\orcidlink{0000-0003-3994-8045}}
\authorrunning{L. Hanu et al.}
%
\institute{Unitary Ltd. \and
University of Amsterdam \and
University of Oxford}
\maketitle
\begin{abstract}
Multi-modal retrieval is an important problem for many
applications, such as recommendation and search. Current benchmarks
and even datasets are often manually constructed and consist of mostly
clean samples where all modalities are well-correlated with the content.
Thus, current video-text retrieval literature largely focuses on video titles
or audio transcripts, while ignoring user comments, since users often tend
to discuss topics only vaguely related to the video. Despite the ubiquity
of user comments online, there is currently no multi-modal representation
learning datasets that includes comments. In this paper, we a) introduce a
new dataset of videos, titles and comments; b) present an attention-based
mechanism that allows the model to learn from sometimes irrelevant data
such as comments; c) show that by using comments, our method is able
to learn better, more contextualised, representations for image, video and
audio representations. Project page: \url{https://unitaryai.github.io/vtc-paper}.
\end{abstract}
    
\section{Introduction}

Training large scale multi-modal models from paired visual/text data from the web has seen great success in video understanding and retrieval. 
However, typically only the caption (i.e. title or ``alt text'') is used, ignoring potentially relevant text present on the web page such as user comments.

We explore how to leverage comments for the task of video-text retrieval. 
We consider how comments can be seen as an extra modality, yet with the peculiar characteristics that they are neither inherently derived from the video (as text from speech or OCR would be), nor are they merely extra captions which can be used in place of the title.
This results in two different, yet equally interesting research questions: 
``Can we use comments to augment and adapt our title representations?'' and ``Can we use them to adapt our video features?'' We address both of these in this paper.

A challenge is that comments may often be only tangentially related to the contents of the video (e.g. ``cool video!''), or may be relevant but non-distinctive (``cute cat!'' applies to many videos). 
Yet, since comments often discuss contextual details lacking from the title or video themselves, we hypothesize that correctly leveraging this signal can improve retrieval, endowing either the query or target features with extra context.

Other modalities can also exhibit this behavior, for example, many current works that learn from audio-visual correspondence~\cite{asano2020labelling,alwassel2020self,morgado2020avid} leverage clean datasets such as Kinetics~\cite{carreira2017quo} or VGGSound~\cite{chen2020vggsound} to learn meaningful correspondence between videos and sound, whereas online videos tend to have for example background music that replaces the actual sounds happening in the video, or images overlaid with sounds. 

In this paper, 
we propose a method that can take advantage of this auxiliary context provided by comments while simultaneously filtering it for meaningful information.
Most current models enforce a strict correlation between the different input modalities under the assumption that all are informative of the content. 
The main intuition of our work is that when training a model on partially unrelated data, we need to introduce a mechanism that allows the model to discount auxiliary data when it is not helpful for the task.

To this end, we build a model with a hierarchical attention structure. 
Current representation learning models that are based on transformer architectures already exploit the idea of an attention mechanism to model the correlation between different \textit{parts} of an input signal. 
For example in text understanding, the attention mechanism is applied per word, allowing the model to understand the structure of natural language. 
Even though in principle one could use the same scheme to model the importance of different text inputs on a per-word basis, we find that this makes it difficult to learn the individual importance of inputs.
Moreover, due to the computational complexity of current transformers (squared with sequence length) this approach would only work for a small number of comments. 
We thus add a second layer of attention \textit{per processed input} that allows the model to assess the amount of information at a higher level of features. 
With quantitative and qualitative experiments, we find that this mechanism aligns well with the intuition that some inputs are very relevant to the problem and others can be disregarded. 

To the best of our knowledge, there is no large scale dataset that contains videos, titles, and user comments. 
Thus, to advance the field of representation learning, we introduce ``\dataset'' (Videos, Titles and Comments), a new dataset of 339k videos and titles with an average of 14 comments per video with which we train and evaluate our representations. 
A more detailed summary of the dataset statistics can be found in the Appendix.
In our experiments we show that we can indeed learn meaningful information from user comments for three different modalities: audio, images, and video and that representations learned can generalize to other datasets.
Additionally, we show that the model can correctly identify whether auxiliary information is informative of the content of a video or not.

\begin{figure}[t]
    \centering
    \includegraphics[width=0.98\linewidth]{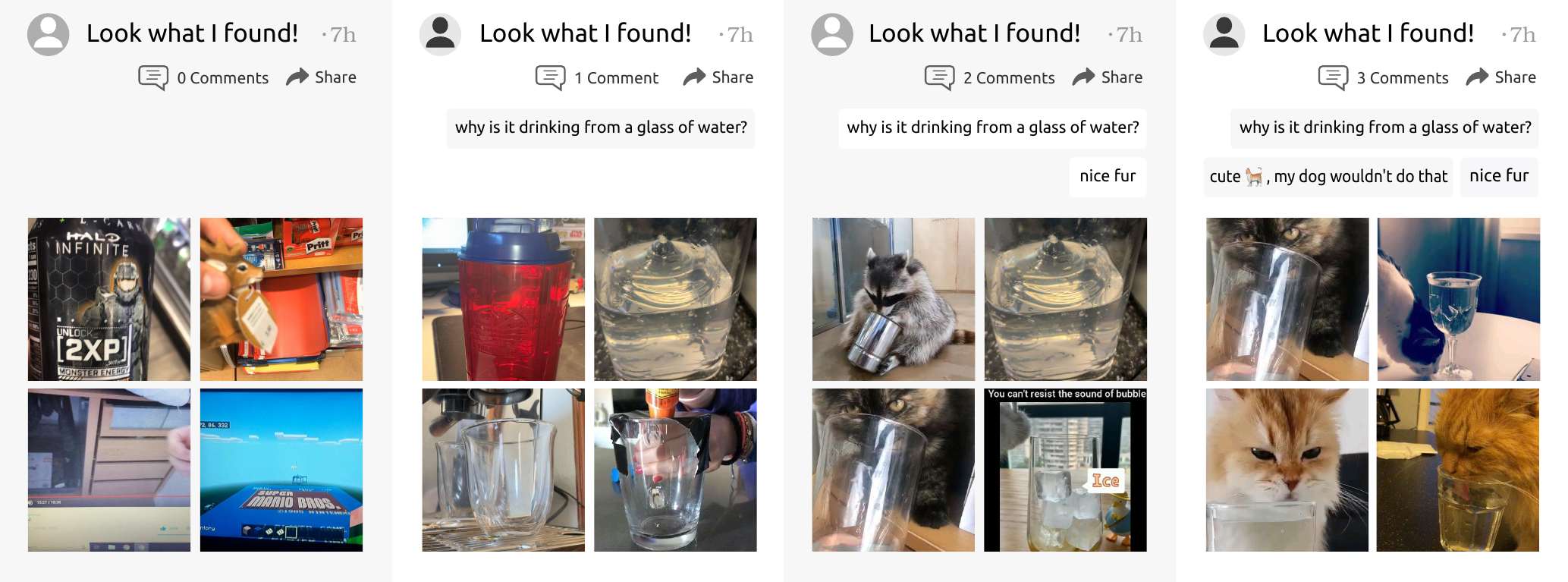}
    \caption{\textbf{Video retrieval from title and comments.} We show the top 4 videos retrieved for the ambiguous title \textit{``Look what I found!}''. From left to right, we progressively add more comments which our model uses to refine the results.}\label{fig:teaser}
\end{figure}

The ability to incorporate auxiliary contextual information also opens up possibilities for useful applications. 
In the video retrieval setting, our method can be used to iteratively refine a text descriptor with new inputs as shown in \cref{fig:teaser}, allowing incremental searching. 
In the zero-shot video classification setting (\ie, ``retrieving'' the correct class description prompt), the prediction for an ambiguous video can be steered towards the correct class using surrounding text from a webpage or user hints.

Overall this paper has three main contributions: 
1) We quantify the value of the \textit{comments} modality for video-text learning.
2) For this, we propose, train, and evaluate on a new dataset \dataset of videos, titles, and comments. 
And 3) we introduce a new hierarchical attention method that learns how to identify relevant auxiliary information, and that can learn representations that even generalize to other datasets.

\section{Related Work}
In this work, we focus on multi-modal learning with a particular focus on learning video-text encoders for retrieval by proposing a novel, multi-modal adaptation module. 

\paragraph{Video-text Pretraining.} Originating from the NLP domain, where the transformer architectures has been a key ingredient and subject to optimization in a multitude of ways~\cite{vaswani2017attention,devlin2019bert,radford2019language,raffel2019exploring,lewis2019bart,clark2020electra,lewis2020pretraining,ruan2022survey,lei2021understanding,lei2021less,li2020hero,fang2021clip2video}, it has recently found applications in the vision-language domain.
For example, recent works have leveraged transformers to learn generalizeable image~\cite{desai2020virtex,sariyildiz2020learning,chen2020generative}, multi-modal image-text~\cite{li2019visualbert,lu2019vilbert,Tan_2019,su2019vlbert,li2020unicoder,chen2019uniter} or video-multilingual text~\cite{huang2021multilingual} representations.
A few works~\cite{Sun_2019,sun2019learning,zhu2020actbert,luo2020univl} combine visual and text modalities as inputs to a BERT model to simultaneously learn semantic video and text representations.
For representation learning, the availability of large-scale datasets such as HowTo100M~\cite{miech2019howto100m} has enabled more effective pretraining of video-text representations for multiple downstream tasks.
More recently, \cite{patrick2020support} show that adding a generative objective to contrastive pretraining can yield gains in video-text downstream tasks. 
Based on the CLIP model~\cite{radford2021learning}, which works well even without finetuning for some retrieval tasks~\cite{luo2021clip4clip},  \cite{bain2021frozen} train video-text CLIP-initialized models by gradually scaling up video training from image training and a custom dataset.
While we also start with a CLIP initialization as in \cite{luo2021clip4clip}, the focus of our paper lies in developing a novel method for leveraging user comments, a modality that has previously been overlooked as a valuable source of information in the text-video retrieval literature.
We also note that there has been a surge in recent vision-text pretrained models inspired by CLIP~\cite{bain2021frozen,mu2021slip,yao2022filip,li2022blip}. 
As we show in the experiments section, our method is agnostic to the pretraining method employed and generalizes beyond CLIP. There are many existing video-text datasets ~\cite{msvd,vtt,lsmdc,hendricks18emnlp,krishna2017dense} but these do not include comments.

\paragraph{Multi-modal domain adaptation.}
While residual adapters for domain adaptation have been explored for uni-modal models such as CNNs, e.g. in \cite{rebuffi2017learning}, there are no works that translate this concept to the multi-modal domain, where cross-modal learning dominates~\cite{alwassel2020self,asano2020labelling,morgado2020avid}.

\paragraph{Vision-text Pretraining.}
While there is a wealth of image-text datasets that provide images with captions, such as OpenImages~\cite{kuznetsova2020open}, ConceptualCaptions~\cite{sharma2018conceptual}, or COCO~\cite{lin2014microsoft}, the recent state of the art methods train on large-scale weakly-supervised datasets that are obtained from image descriptions from for example Reddit (RedCaps~\cite{desai2021redcaps}) or YFCC~\cite{thomee2016yfcc100m}.

\paragraph{Comment Datasets.} To the best of our knowledge, there is only one vision dataset which does include user-comments, the LiveBot dataset~\cite{livebot}.
However this dataset, which contains under 3000 videos, is constructed for artificial comment generation and uses the less common ``video barrage'' (\textit{i.e.} time-synchronous) type of comments.
Despite this, we evaluate our method on this dataset and also find performance gains for video-text representation learning when using comments.
In the context of learning from comments, there is little prior work. 
While the work of~\cite{halevy2020preserving} is somewhat related, as it uses comments and reactions to posts to refine harm predictions on a social media site,  we are the first to demonstrate that user comments can be used as a complementary modality when learning video-text representations.

\section{\dataset Dataset}

We collect a dataset ``\dataset'' of videos along with their titles and comment threads from social news site reddit.com, using their provided API. 
The videos are collected and used in a manner compatible with national regulations on usage of data for research.
Unlike most curated video datasets, this data is more representative of the types of videos shared ``in the wild'', containing a large proportion of videogames, screenshots and memes.

Using a classifier trained on a small amount of labelled data, we estimate that videogame footage makes up 25\% of examples, other screenshots, memes and comics make up 24\%, live action footage is 49\% and artistic styled content (such as drawn animation) is 2\%. 
The average video length is 33s.

From 1 million raw videos collected, we perform deduplication and filtering, ending up with a training set of 461k videos. For the experiment in \cref{tab:noface} on training without faces we do further filtering to remove faces, finding that about 65\% of videos contain a face.
To compensate for the decrease in quantity of training data we gather extra non-face-containing videos, ending up with 339k videos.
For the evaluation results we use a test set consisting of 5000 videos with at least three comments each.

For each example in the dataset we obtain:
The \texttt{title} of the post, a high quality \texttt{preview image}, which is generated automatically by Reddit, typically 640 pixels wide and corresponding to the middle frame of the video, the \texttt{video} itself, downloaded in low quality and resized to have height 320 pixels for storage reasons, and up to 500 randomly selected \texttt{comments} per post.

For all the image-based experiments, we use the high quality preview image, whereas for the video experiments in \cref{tab:video} we use the extracted video frames. 
To fairly compare video and image models given the lower video resolution and quality, in \cref{tab:video} the ``1 frame'' case corresponds to the first frame from the video rather than the high quality preview image.

\paragraph{Deduplication}

We use the GPU implementation of the FAISS similarity search toolkit \cite{JDH17} to efficiently deduplicate the dataset by indexing the video thumbnail embeddings obtained from a ImageNet pretrained ResNet18~\cite{he2016deep}.
These indices are then used to discard video entries with a high similarity to other posts.

\paragraph{Safety and Privacy}
Additionally, we remove toxic text content (such as slurs and hate speech) from titles and comments using the detoxify library \cite{Detoxify}. 
\Cref{tab:detoxifyres} show the prevalence of content that has been removed this way. 

\begin{SCtable}
\footnotesize
\centering
\setlength{\tabcolsep}{2pt}
\captionof{table}{\textbf{Prevalence of toxic text before filtering.} We report the proportion of posts, titles, and comments that are flagged as having potentially offensive content by the open-source library Detoxify. We use a threshold of $0.9$}
\begin{tabular}{lrrr}
\toprule
 \textbf{Detoxify label} &  \textbf{\% titles} &  \textbf{\% comments} \\
\midrule
   toxicity           &      2.32 &        5.62 \\
   severe toxicity    &      0.00 &        0.00 \\
   obscenity          &      1.23 &        3.73 \\
   identity attack    &      0.00 &        0.00 \\
   insult             &      0.82 &        1.95 \\
   threat             &      0.05 &        0.07 \\
   sexually explicit  &      0.09 &        0.22 \\
\bottomrule
\end{tabular}
\label{tab:detoxifyres}
\end{SCtable}
It is crucial that a dataset is well-conceived and potential risks are thought-out before release.
We take two steps to ensure safety and usefulness of our proposed dataset. 
First, for the releasing the dataset we further filter the dataset to exclude videos that contain faces using the automatic face-detection filtering process from PASS~\cite{asano2021pass}.
In our experiments we show that this does not lead to a significant change in performance.
Second, we provide a \textit{Datasheet}~\cite{gebru2021datasheets} for the proposed dataset which can be found in the supplementary material. 
This dataset will be released for research use together with the paper.
\section{Methods}

In this section we will first recap the mechanism behind current contrastive, multi-modal representation learning methods that rely on clean data.
We will then introduce our Context Adapter Module that allows learning from the auxiliary modality through an attention mechanism.
Finally, we will describe how we can extend an existing backbone for images to videos and audio, to be able to leverage large, pretrained models.

\subsection{Background}
In multi-modal representation learning we are given a dataset $\mathcal{X}$ of $N$ samples $x_i \in \mathcal{X}, i \in \{1, \ldots N\}$ that individually consists of different signals.
Most previous work focuses on two modalities and we will---for now---also adhere to this standard to simplify the notation. 
This means that each input sample $x_i = (v_i, t_i)$ is a pair of---in our case---a visual input $v_i \in \mathcal{V}$ and its associated text, often the title, $t_i \in \mathcal{T}, 1 \leq i \leq N$.

The goal is now to learn mappings $f_v: \mathcal{V} \mapsto \mathcal{Y}, f_v(v_i) = \phi_{v,i}$ and $f_t: \mathcal{T} \mapsto \mathcal{Y}, f_t(t_i) = \phi_{t,i}$ from each of the modalities to a $d$-dimensional, joint embedding space $\mathcal{Y} = \mathbb{R}^d$. 
Recent methods, such as~\cite{radford2021learning}, learn the mapping (in their case from images and their captions) to the embedding space with a double contrastive loss over a mini-batch $\mathcal{B} \subset \mathcal{X}$ using an affinity matrix $A$ computed between all pairs of samples in the batch:
\begin{equation}\label{eq:affinity}
    A_{ij} = \left\langle\, \frac{\phi_{v,i}}{\sqrt{\tau} \| \phi_{v,i} \|} , \frac{\phi_{t,j}}{\sqrt{\tau} \| \phi_{t,j} \|} \right\rangle
\end{equation}

An entry $A_{ij}$ measures the similarity between the embeddings $\phi_v(v_i)$ and $\phi_t(t_i)$ via cosine similarity that is scaled by a temperature parameter $\tau$.
The idea is now to maximize the similarity between the embeddings from the \textit{same} sample, \ie the diagonal of $A$ and minimize all non-diagonal entries. 
This can be achieved efficiently using a double-contrastive formulation that operates across columns and rows of $A$,
\begin{equation}
    \mathcal{L}(A) = \frac{1}{2}
    \sum_{i=1}^{|\mathcal{B}|} 
    \frac{A_{ii}}{\log \sum_{j=1}^{|\mathcal{B}|} \exp A_{ij}} +
    \frac{A_{ii}}{\log \sum_{j=1}^{|\mathcal{B}|} \exp A_{ji}}.
\end{equation}

This formulation has the neat effect that it accomplishes maximizing the diagonal entries and minimizing all other entries of $A$ in one self-balancing formulation.
However, it makes the critical assumption that both modalities are equally informative of each other.
In the case of sometimes irrelevant data, or when one modality has much less information content than the other (\eg ``\textit{nice video!}''), this assumption does not hold and training with this objective will result in a very volatile learning objective and thus a sub-optimal joint embedding.

In the next section we will introduce our Context Adapter Module that is able to deal with this type of inputs by allowing it to discount information when it is not relevant for the context.

\subsection{Context Adapter Module}

\begin{figure}
\centering
\begin{subfigure}{.5\textwidth}
  \centering
  \includegraphics[width=0.95\linewidth]{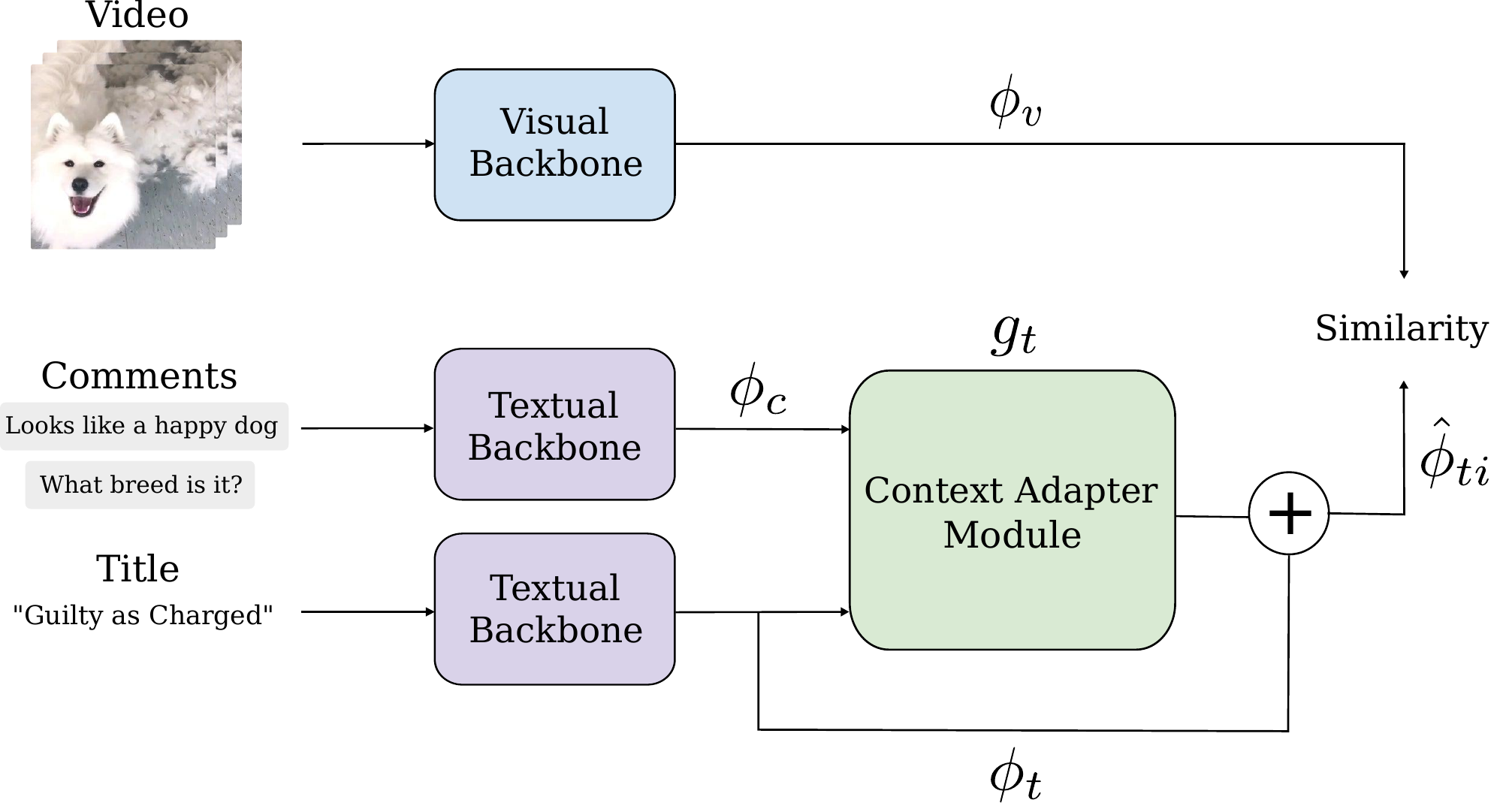}
  \caption{Adapting the text branch.}
  \label{fig:sub1}
\end{subfigure}%
\begin{subfigure}{.5\textwidth}
  \centering
  \includegraphics[width=0.95\linewidth]{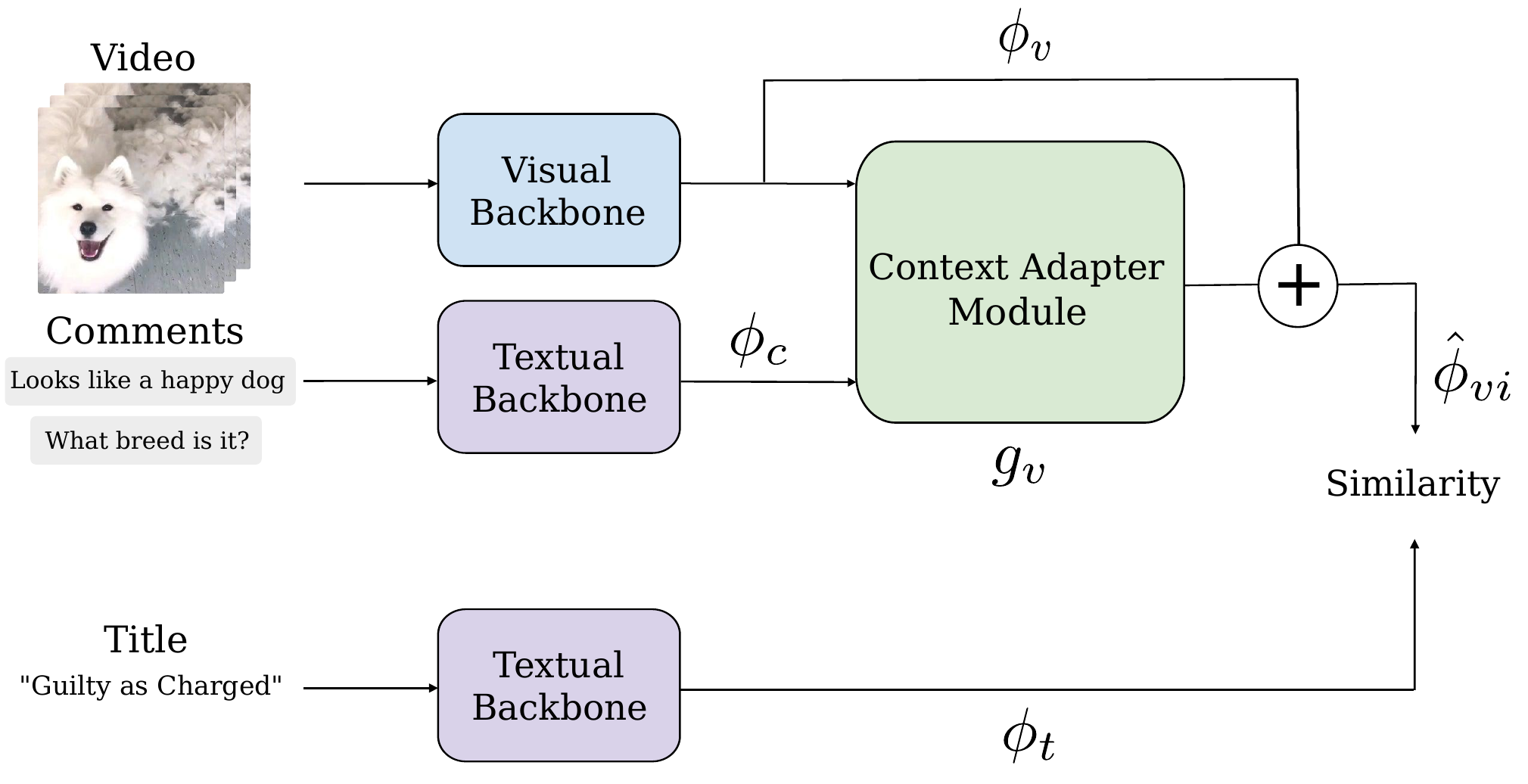}
  \caption{Adapting the visual branch.}
  \label{fig:sub2}
\end{subfigure}
\caption{\textbf{Method Overview.} We introduce a context adapter module that uses inputs of the auxiliary modality to adapt the embedding of another branch. With this module the model is able to accept or discount information.}
\label{fig:method}
\end{figure}

In order to capture and filter the relevant information from the comments, we propose a transformer-based Context Adapter Module (CAM) which operates in a residual fashion, additively adapting either the visual or text branch of CLIP with contextual information obtained from the comments (see~\cref{fig:method}). 
Formally, we are now adding another modality---the comments---to the input which extends it to $x_i = (v_i, t_i, c_{i,1}, \ldots, c_{i,M})$ with $c_{i,k} \in \mathcal{T}$. To reduce clutter in the notation, we have defined a fixed number of comments $M$ for each sample.

Since both title and comments share the same modality (\ie text), we can leverage the same encoder to transform comments to embeddings $f_t(c_{ik}) = \phi_{c,ik}$.

As we expect the comments to be sometimes unrelated, our Context Adapter Module needs a mechanism to discount off-topic comments and update the primary modality $\phi_v(v_i)$ or $\phi_t(t_i)$, steering it in the most informative direction.

We introduce this mechanism as a function of both the primary modality and the comment embeddings $\phi_{c,ik}$, as we want to compare
the informativeness of all these inputs at a high level.
To this end, we design adapter modules $g_v$ and $g_t$ that extract information from the comments in the form of a residual:
\begin{equation}
    \hat{\phi}_{oi} = \phi_{oi} + g_o(\phi_{oi},\, \phi_{c,i,1},\, \ldots,\, \phi_{c,i,M}) \, , \, o \in \{v, t\}
\end{equation}

With the adapted embeddings $\hat{\phi}_{vi}$ and $\hat{\phi}_{ti}$ we recompute the affinity matrix (now $\hat{A}$) (Eq.~\ref{eq:affinity}) and use it for the loss $\mathcal{L}(\hat{A})$.
This design has several advantages. 
On one hand, extracting ``only'' a residual from the auxiliary inputs $c_{ik}$ means that the model is easily able to ignore them by predicting $g(\cdot) = 0$.
On the other hand, this effectively allows us to skip the adapter module when we evaluate without comments, while still learning the joint embedding from richer data.
 
In practice, we implement $g$ as a small transformer architecture.
Rather than operating on tokenised words, this transformer operates on embeddings ($\phi_{vi}$ and $\phi_{ti}$) themselves, taking as input the encoded feature from the branch to be adapted, along with comment features $\phi_{c,ik}$.
By treating embeddings as tokens in their own right, we allow the embeddings to attend to each other and learn what combinations of the inputs should be used to update the original feature through the residual connection. 
 
Additionally, to avoid bleeding information between the two modalities through the Context Adapter Module, during training, we only adapt either the video embedding with $g_v$ or the text embedding with $g_t$. 
If we would use both adapters simultaneously, there is a trivial solution that minimizes the loss $\mathcal{L}$: when the adapters learn to remove the original embedding through the residual, \eg $g_o(\phi_{oi},\, \{\phi_{c,i,k},\}) = -\phi_{oi} + \phi_{c,i,1}$ both adapted embeddings become the same $\hat{\phi}_{vi} = \hat{\phi}_{ti}$ which trivially maximizes their similarity, thus preventing the model to learn a meaningful modality alignment.
To prevent the model from learning a transformation of the embedding space through the residual, we train only one adapter at a time. We also randomly skip the adapter entirely with probability 0.5, which ensures that the un-adapted features are still
meaningful in isolation, and the adapter can be bypassed at evaluation time if comments are not available.

\subsection{Video}\label{sec:timesformer}
To leverage the capacity of large pre-trained computer vision models, we adopt the architecture by~\cite{radford2021learning} as our backbone models $f_v$ and $f_t$.
While this transformer was trained on a huge volume of image-text data, it cannot be applied directly to videos since it is built for images and has no temporal extent.
To take advantage of the temporal information present in video data, we use the Divided Space-Time attention mechanism recently introduced in the
TimeSformer architecture~\cite{bertasius2021space}. 
We modify the image transformer architecture by adding patchwise self-attention across 8 frames in time to each of the 12 residual
attention blocks, followed in each case by a zero-initialised linear layer.
We also add a learned temporal position embedding which is summed to the input and again zero-initialised.
The initialisation is transparent, such that when loading pretrained weights trained from images, at initialization time, the modifications do not affect the inference of the model.
During training, the model can then gradually activate the additional temporal components to learn from the temporal information of a video.
Full details on the architecture are provided in the Appendix.

\subsection{Audio}
To further compare the effect of the newly proposed comments modality with another common modality besides text, we also conduct experiments using audio. 
For this we utilize the audio-encoder from GDT~\cite{patrick2020multimodal} that was pretrained on a large video-audio dataset.
The audio-encoder works on 2s audio segments converted into a spectrogram, please see the Appendix for further details.
\section{Experiments}
This section has two main objectives. The first is to show how the additional modality of user comments can be used to improve multi-modal representation learning.
Second, the experiments show how our new dataset \dataset can be used to learn video, audio, image and text representations. 

\paragraph{Implementation details.}\label{sec:impl-details}

We use CLIP \cite{radford2021learning} (ViT-B/32 checkpoint unless otherwise mentioned) as the initialisation for the backbone. 
Our concrete implementation of the CAM $g$ is a 2-layer transformer, consisting of two residual multihead self-attention blocks. The input consists of $M+1$ input embeddings (for the $M$ comments and title/video embedding $\phi_{oi}$) having 512 dimensions each.
Each block performs 8-head self-attention on the inputs, followed by two linear layers with output size 2048 and 512 respectively. LayerNorm normalisation is used, along with GELU activation following the first linear layer.
From the $M+1$ outputs of the transformer, we then normalize, take the mean and renormalize.
We use the Adam~\cite{kingma2014adam} optimizer with a learning rate of $1\times10^{-6}$ when training the entire model on its own or with the adapter.
All implementation and architecture details can be found in the Appendix.

We report the standard Recall@N metrics as a percentage (often abbreviated R@N), giving the proportion of results
where the ground truth is ranked in the top N. We show both Text-Video-Retrieval (TVR) and Video-Text-Retrieval (TVR). Unless otherwise mentioned we use 5 comments for evaluation.

\subsection{Additional Datasets}\label{sec:dataset}
\paragraph{LiveBot Dataset.}

Prior work on building a dataset with videos and comments is LiveBot \cite{livebot}, which consists of 2361 videos 
and 895,929 comments, obtained from Chinese social network Bilibili. This differs a lot from our setting, since the comments
in question are made while the video is being streamed live and associated with certain timecodes, and comments and titles are in Chinese rather than English. Nevertheless, in order to evaluate how well our method works for this
sort of data, we use automatic translation to translate the titles and five comments for the 100 videos in the LiveBot test set, which we call LiveBotEN and show in \cref{tab:video}. Due to duplicate video and missing split metadata in the original LiveBot release, we follow the split used in \cite{wu2020response}.

\paragraph{KineticsComments.}
As an additional video dataset with comments, we construct a dataset based on Kinetics-700 \cite{kinetics,carreira2019short}, for which we download the videos along with associated YouTube metadata including title, description and comments.
We translate non-English titles and descriptions into English using a commercial translation API. We use the title as the primary text modality, and for auxiliary context we use comments. We construct a test set, consisting of videos from the Kinetics test set for which we have at least 3 comments, giving a set of 6292 videos which we use to evaluate our method in~\cref{tab:video}.

\subsection{Evaluating the Context Adapter Module}
In this section we evaluate our Context Adapter Module on the above described datasets with comments. 

\paragraph{Context Adapter Module.}
\begin{table}[htb]
    \footnotesize
    \centering
    \setlength{\tabcolsep}{4pt}
    \captionof{table}{\textbf{Adaption Mechanisms.} Comparing different ways in incorporate auxiliary information: adapting the title with 5 comments}
    \begin{tabular}[t]{@{}lrrrr@{}}
        \toprule
        \textbf{Method}  & \textbf{TVR R@1}  & \textbf{TVR R@10}  & \textbf{VTR R@1} & \textbf{VTR R@10} \\ \midrule
        no comments (zero-shot)         & 11.1 & 26.0 & 11.1 & 25.3 \\
        no comments (fine-tuned)        & 15.5 & 34.9 & 14.4 & 33.4 \\ 
        averaging (zero-shot)           &  7.3    & 22.7     &   6.9  &   20.0   \\
        averaging (fine-tuned)          & 16.6 & 42.3 & 18.1 & 43.3 \\ \midrule
        \textbf{ours}        & \textbf{18.4} & \textbf{43.2} & \textbf{18.6} & \textbf{44.0} \\
        \bottomrule
    \end{tabular}
\label{tab:adaptiontype}
\end{table}

To verify that the Context Adapter Module is indeed able to learn better representations from the comment modality, we compare it to several baselines in \cref{tab:adaptiontype}.
The most trivial baseline is to ignore any comments and to train simply with image-title pairs. 
This results in the lowest performance, showing that there is valuable information in the comment data.
Another baseline consists of averaging the features from the titles with the features of the comments, which is a direct way to incorporate the comments. 
We make these baselines stronger by fine-tuning the backbone during training which does result in a performance improvement.
Finally, a baseline where all text is concatenated would be interesting to evaluate, however due to memory/text-length limitations concatenating more than 2-3 comments is intractable with current encoder architectures. 

Finally, our context adapter module is able to improve over all baselines. 
We hypothesize that this comes from the ability of the adapter module to ignore irrelevant comments. 
To test this, we perform an experiment where we add random irrelevant distractor comments (during evaluation only) and measure the impact of distractors on the performance. 

\begin{SCfigure}[1.0][t]
    \centering
    \includegraphics[height=0.25\textheight]{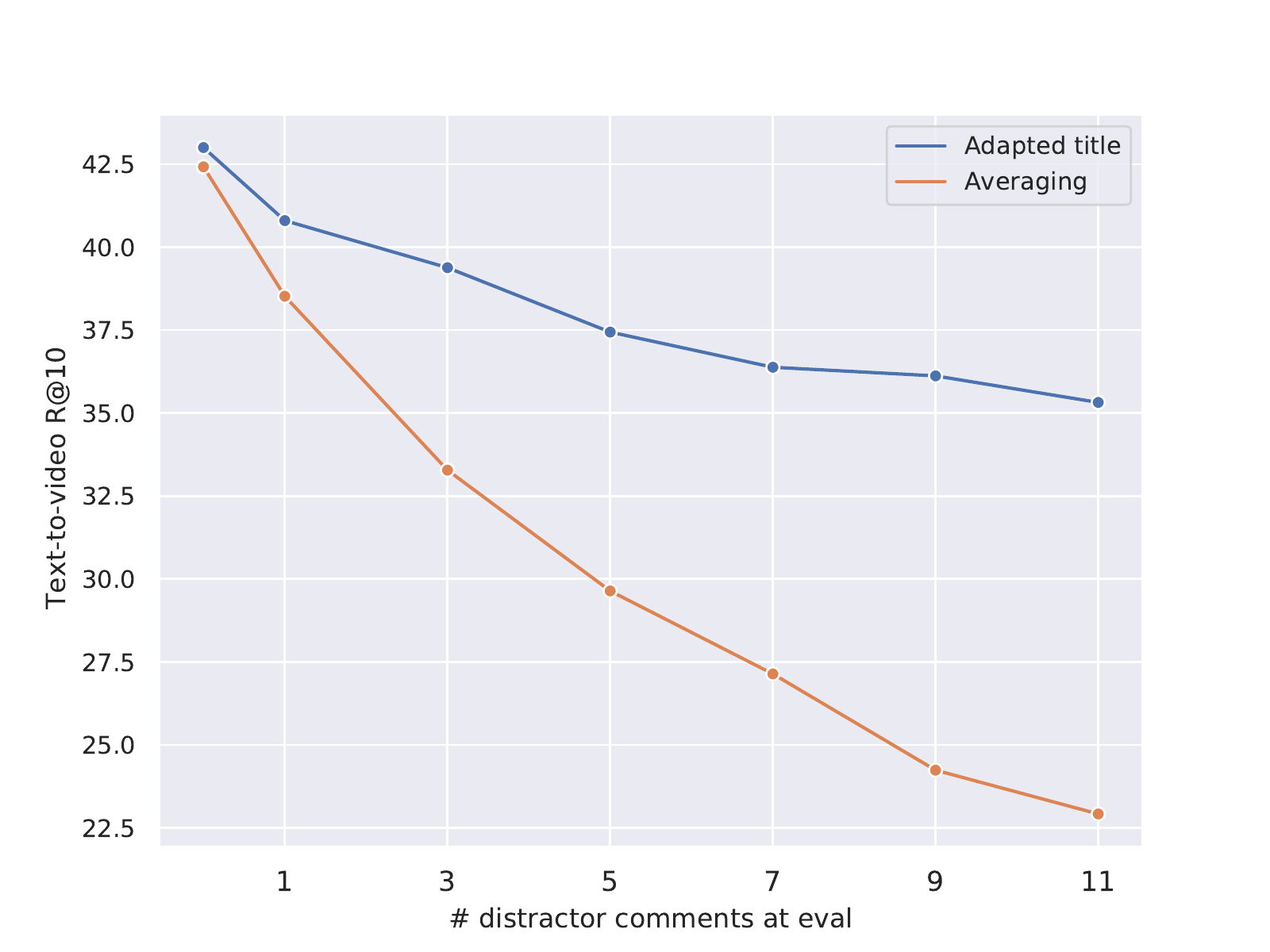}
    \caption{\textbf{Influence of Distractor Comments.} We gradually add irrelevant distractor comments during evaluation. The context adapter module is able to deal with irrelevant information much better than baseline, showing that it has learned to down-weigh uninformative content }
    \label{fig:distractors}
\end{SCfigure}

The results of this experiment can be seen in \cref{fig:distractors}, where we gradually increase the number of distractors and evaluate retrieval performance. 
The averaging baseline is strongly affected by this ``misinformation'' whereas the context adapter module has implicitly learned to ignore irrelevant information during training. 
Note that there is no explicit supervision for this during training and the model has to learn this behavior directly form the data.
As the backbone is trained also for the averaging baseline, both methods can learn to ignore generally uninformative content (``look what I found'') but the context adapter module can learn to exploit the context of the title with relation to the comments through the attention mechanism. 

\paragraph{Comparing Encoders.}

\begin{table}[t]
    \footnotesize
    \centering
    \setlength{\tabcolsep}{4pt}
    \captionof{table}{\textbf{Backend Fine-tuning.} Effect of fine-tuning the encoders}
    \begin{tabular}[t]{@{}lrrrr@{}}
        \toprule
        \textbf{Method}  & \textbf{TVR R@1}  & \textbf{TVR R@10}  & \textbf{VTR R@1} & \textbf{VTR R@10} \\ 
        \midrule
        no fine-tuning     & 11.1 & 26.0 & 11.1 & 25.3 \\
        fine-tuning        & 15.5 & 34.9 & 14.4 & 33.4 \\
        \bottomrule
    \end{tabular}
\label{tab:finetuning}
\end{table}

As in all current multi-modal approaches, the architecture and pre-training of the visual/text/audio encoder is important.
In \cref{tab:finetuning} we show that fine-tuning the (in this case CLIP \cite{radford2021learning}) encoder does improve the performance by a significant margin. 
This shows that even though the encoder was trained on an extremely large image/text dataset, there is a domain gap with \dataset (videos and comments) that can be bridged by fine-tuning.

\begin{table}[t]
    \footnotesize
    \centering
    \setlength{\tabcolsep}{4pt}
    \captionof{table}{\textbf{Encoder Backbones.} Comparing different pre-trained encoders. We keep the encoders frozen and just train the CAM. Showing Recall @ 10, retrieving image from text+comments}
    \begin{tabular}[t]{@{}lccc@{}}
        \toprule
        \textbf{Backbone} & No Comments & 5 comments & 20 comments \\ 
        \midrule
        FiT             \cite{bain2021frozen}         & 8.8   & 12.0  & 12.8  \\
        SLIP (ViT-B)    \cite{mu2021slip}             & 9.3   &   10.2   &   11.6   \\
        CLIP (ResNet50) \cite{radford2021learning}    & 22.7  &   27.4   &   27.9   \\
        CLIP (ViT-B/32) \cite{radford2021learning}    & 25.3  &  32.3    &   34.1   \\
        CLIP (ViT-L/14) \cite{radford2021learning}    & 32.9  & 42.0  &  44.1  \\
        \bottomrule
    \end{tabular}
\label{tab:backbones}
\end{table}

In \cref{tab:backbones} we compare different model types of CLIP \cite{radford2021learning} with other current models: SLIP \cite{mu2021slip} and FiT \cite{bain2021frozen}. 
Naturally, larger architectures perform better, in line with ResNet50 falling behind ViT based encoders for CLIP.
Comparing to CLIP, FiT and SLIP have been trained on roughly two orders of magnitude smaller datasets (400M image-text pairs for CLIP) resulting in decreased image and text understand capabilities. 

\begin{SCfigure}[1.0][t]
    \centering
    \includegraphics[height=0.25\textheight]{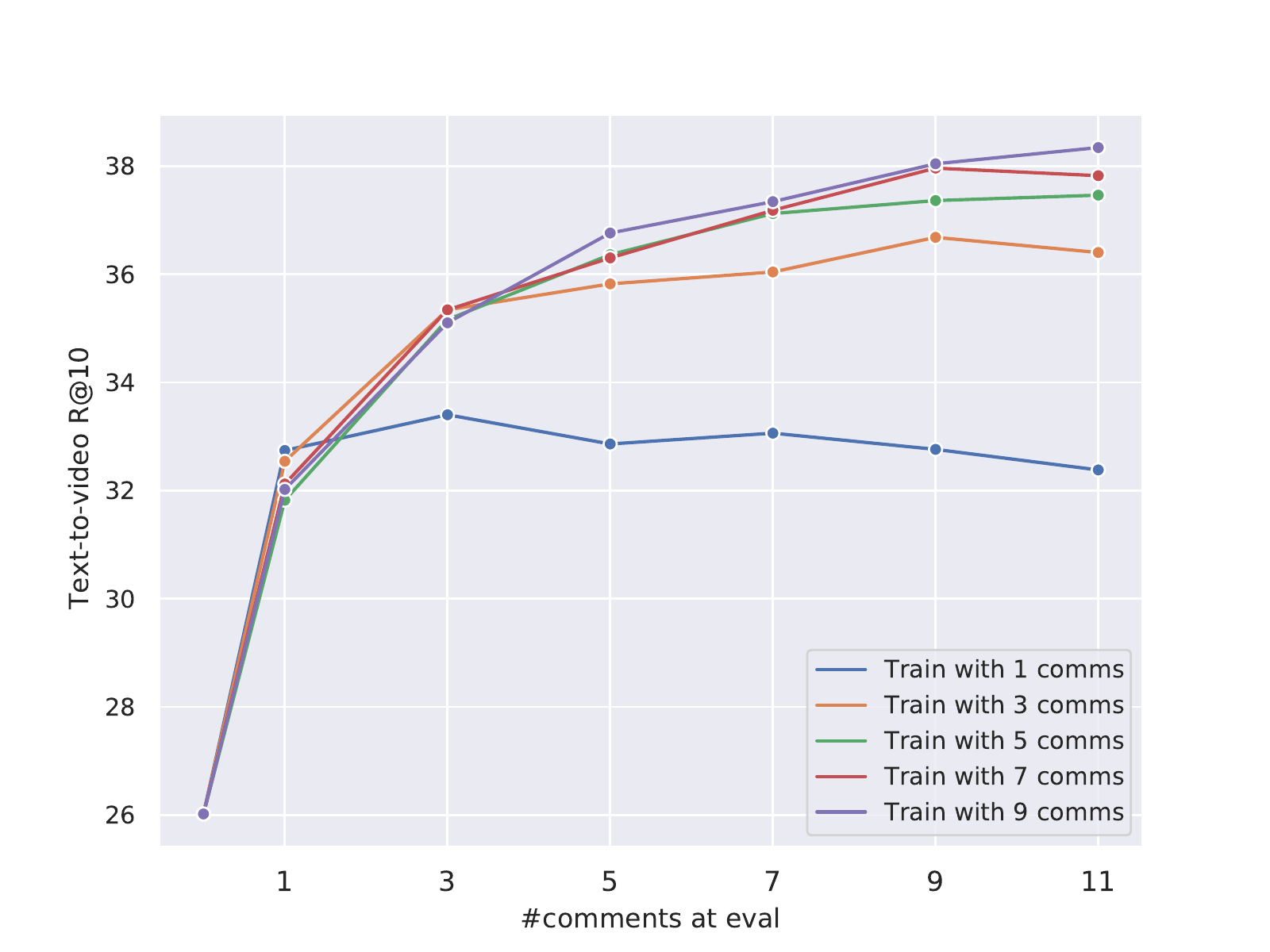}
    \caption{\textbf{Varying Number of Comments.} We show the influence of varying the number of comments during training and testing time. All variants benefit from using comments. Training with a single comment is not enough to learn a stable filtering behavior}
    \label{fig:varyingcomments}
\end{SCfigure}

Additionally, we find that adding comments improves the performance of \textit{all} encoders. 
Adding more comments consistently improves the performance again, however with diminishing returns.
We further investigate this behavior in \cref{fig:varyingcomments}, where we vary the number of comments during training and evaluation time.
All models benefit from using comments compared to not using comments.
Interestingly, training with one comment seems to be insufficient to learn how to extract additional information when there is more than one available. 

\paragraph{Different Modalities.}
\begin{table}[t]
    \footnotesize
    \centering
    \setlength{\tabcolsep}{4pt}
    \captionof{table}{\textbf{Adaption Modality.} Comparing different ways to incorporate auxiliary information: adapting the title or image with comments}
    \begin{tabular}[t]{@{}lrrrr@{}}
        \toprule
        \textbf{Method}  & \textbf{TVR R@1}  & \textbf{TVR R@10}  & \textbf{VTR R@1} & \textbf{VTR R@10} \\ 
        \midrule
        none         & 15.5 & 34.9 & 14.4 & 33.4 \\
        title        & 18.0 &43.2 &18.7& 43.9 \\
        image        & 28.2 & 51.2 & 25.1 & 49.9  \\
        \bottomrule
    \end{tabular}
\label{tab:adaptionmodality}
\end{table}

The intuition behind the context adapter module is that it allows to adapt information in a feature with potentially unreliable auxiliary data. 
As described earlier, the comments can be used to either adapt the information in the image or in the title. 
In \cref{tab:adaptionmodality} we compare these two options and find that adapting the image results in a larger performance improvement than adapting the title. 
This can be explained by the modality gap between visual and textual information. 
When adapting the image information with the text from the comments, the context adapter module can learn to close the information gap between text and image much more effectively than when adapting the tile with the text from the comments.
However, we find that in the context of retrieval and multi-modal representation learning a more realistic (and challenging) scenario is posed when the title is adapted (as for example seen in \cref{fig:teaser}). 

\begin{table}[htb]
    \footnotesize
    \centering
    \setlength{\tabcolsep}{4pt}
    \captionof{table}{\textbf{Combining Modalities.} We show that our method is robust to different combinations of modalities, both at train and at test time}
    \begin{tabular}[t]{@{}llrrrr@{}}
        \toprule
        &  &   \multicolumn{2}{c}{\textbf{Text $\rightarrow$ Video}} & \multicolumn{2}{c}{\textbf{Video $\rightarrow$ Text}} \\

        \textbf{training} & \textbf{inference} & \textbf{ R@1}  & \textbf{ R@10}  & \textbf{ R@1} & \textbf{ R@10} \\ 
        \midrule
        CLIP & img+title  & 11.1 & 26.0 & 11.1 & 25.3 \\
        img+title      & img+title          & 15.5 & 34.9 & 14.4 & 33.4 \\
        img+title+cmts & img+title          & 15.5 & 34.5 & 14.4 & 33.3 \\
        img+title+cmts & img+title+cmts     & 18.0 & 43.2 & 18.7 & 43.9 \\ \midrule
        img+title+cmts+audio        &     img+title               &    15.4  &    34.0  &   14.3   &   32.9   \\
        img+title+cmts+audio        &     img+title+audio         &    15.8  &    36.9  &   12.2   &   30.4   \\
        img+title+cmts+audio   &  img+title+cmts+audio    &    19.6                & 45.6     &    20.6  &   47.2      \\
        \bottomrule
    \end{tabular}
\label{tab:traintest}
\end{table}

Another benefit of the context adapter module is that, not only can it deal with a variable number of comments during inference, but it also allows for evaluation without any comments. 
\Cref{tab:traintest} shows that training with comments does not have an impact on the performance of the model in a setting where no comments are available at test time. 
This means that the learned model is flexible and can be used in both settings directly and without any changes. 

The idea of learning from potentially unreliable auxiliary data extends beyond the use of comments and in \cref{tab:traintest} we perform additional experiments using the audio information in the videos. 
In many current video datasets the quality of the audio varies drastically. 
For example, some videos replace the natural audio with music, removing any aural clues about the content of the video.
Similar to comments, including audio in the context adapter module during training allows the model to identify irrelevant audio information.
This results in a virtually unchanged performance when no audio is available during test time, but further improves the final performance when considering all four modalities.

\paragraph{Video Data.}
In this section we evaluate the impact of using videos instead of single frames in combination with also adding comments. For video evaluation we take the 8 initial frames with a stride of 16.
In \cref{tab:video} we find that on all datasets adding comments boosts the retrieval performance for both video-to-text (VTR) and text-to-video (TVR) significantly, confirming the value of the modality.
It is important to note, that all models were trained only on \dataset and the improvements translate directly to KineticsComments and LiveBotEN.
In most cases, the improvement gained from adding comments is considerably larger than the information gained by incorporating temporal information. 
This is an additional data point for the importance of the comment modality.
While \dataset test set does not benefit largely from video training itself, using videos during training still improves the performance on the other datasets. 

\begin{table}[htb]
    \footnotesize
    \centering
    \setlength{\tabcolsep}{6pt}
    \captionof{table}{\textbf{Video results.} Experiments using video frames. Trained adapting the video branch with comments, with either one or eight frames from the video. Showing Recall@10}
    \begin{tabular}[t]{@{}llllllll@{}}
        \toprule
        & &  \multicolumn{2}{c}{\textbf{\dataset}} & \multicolumn{2}{c}{\textbf{KineticsComms}} & \multicolumn{2}{c}{\textbf{LiveBotEN}} \\
        \textbf{inference} & \textbf{\#frames}  & \textbf{VTR}  &  \textbf{TVR} & \textbf{VTR}  &  \textbf{TVR}  & \textbf{VTR}  &  \textbf{TVR} \\ 
        \midrule
        video           &    1  &  28.9  &  28.3 &   48.8 & 46.9  & 48.0 & 49.0  \\
        video+comments  &    1  &  40.8  &  41.0 &   61.1 & 59.2  & 64.0 & 64.0  \\
        \midrule
        mean-pooling    &    8  & 19.3    & 24.2  &  54.1 & 49.8  & 69.0 & 66.0   \\
        video           &    8  & 28.9    & 27.6  &  56.9 & 55.8  & 70.0 & 72.0 \\
        video+comments  &    8  & 41.5    & 41.9  &  68.0 & 66.1  & 69.0 & 80.0 \\
        \bottomrule
    \end{tabular}
\label{tab:video}
\end{table}

\paragraph{Privacy.}
Finally, we perform an experiment on the effect of removing all videos from the dataset that contain a face.
\Cref{tab:noface} shows that even though this reduces the size of the training set, the performance is not negatively affected.
The evaluation is performed on the same test set. 
We can even see a small increase in performance, that could potentially be attributed to a more balanced training set, as videos of humans tend to dominate the dataset before the face removal.

\begin{table}[t]
    \footnotesize
    \centering
    \setlength{\tabcolsep}{4pt}
    \captionof{table}{\textbf{Privacy -- Removing Faces.} Effect of removing all videos that contain a face from the dataset. The evaluation is performed on the same test set (that contains faces). The difference in performance is marginal}
    \begin{tabular}[t]{@{}lrrrr@{}}
        \toprule
        \textbf{Method}  & \textbf{TVR R@1}  & \textbf{TVR R@10}  & \textbf{VTR R@1} & \textbf{VTR R@10} \\ 
        \midrule
        with faces     & 18.0 & 43.2 & 18.7 & 43.9 \\
        without faces  & 18.2 & 44.0 & 18.1 & 45.0 \\
        \bottomrule
    \end{tabular}
\label{tab:noface}
\end{table}

\section{Discussion and Conclusions}

\paragraph{Limitations.}\label{sec:limitations}

\begin{SCfigure}[1.0][t]

    \includegraphics[width=0.45\textwidth]{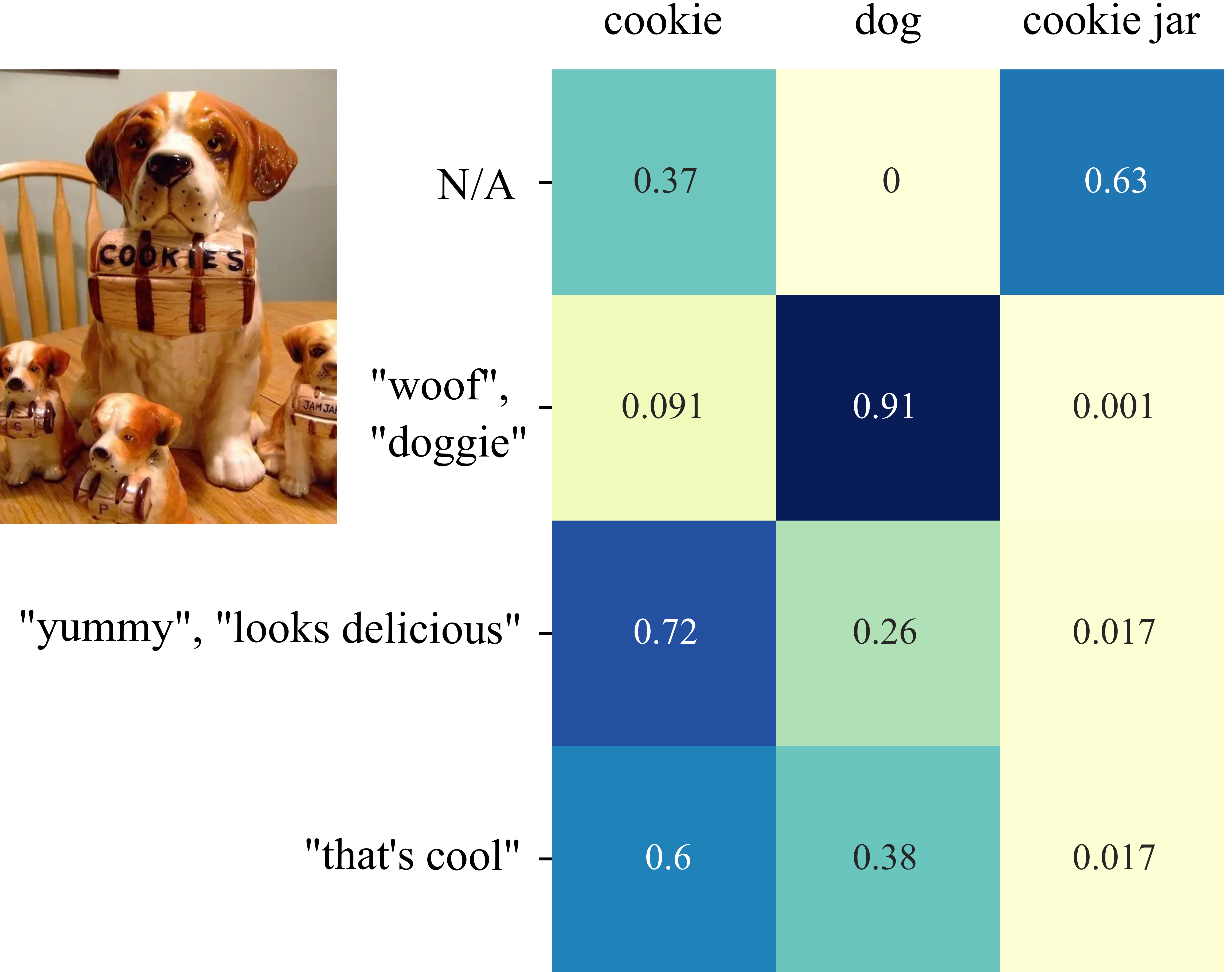}
    \caption{\textbf{Failure Case.} A heatmap showing the similarities between the image adapted with different comments (rows), and captions (columns). The adapter can steer away the embedding from the right association ``cookie jar'' depending on the comment input. This means that adversarial comments could affect the performance of the model}
    \label{fig:heatmap}
\end{SCfigure}

We find that the context adapter can be led to override the information in a title if we adversarially craft comments that all point to different content.  
Qualitative examples of this can be seen in \cref{fig:heatmap}. 
The model without comments, correctly associates the image with a cookie (jar), however when adding a comment about a ``dog'' the model prefers the dog label over cookie.

\paragraph{Conclusion.}
We have presented \dataset, a new dataset with videos, titles, and comments and a context adapter module, which is able to extract information from auxiliary input sources for learning a joint, multi-modal embedding. 
The dataset fills a gap in current vision-text datasets as it includes comments that potentially provide additional information about the content.
In our experiments, we are able to show that learning from comments improves video-text retrieval when adapting the representation with user comments. Moreover, the context adapter module is able to identify whether an auxiliary input is relevant to the content in the other modalities or not. 
This mechanism could, for example, be used to filter datasets for meaningful auxiliary content.

\paragraph{Acknowledgements.}
This project is supported by Innovate UK (project 71653) on behalf of UK Research and Innovation (UKRI). Y.M.A. and C.R. were also supported by an AWS Machine Learning Research Award (MLRA). We also thank Sasha Haco from Unitary for her support for this project.

\clearpage

\bibliographystyle{splncs04}
\bibliography{refs}

\begin{thebibliography}{10}
\providecommand{\url}[1]{\texttt{#1}}
\providecommand{\urlprefix}{URL }
\providecommand{\doi}[1]{https://doi.org/#1}

\bibitem{alwassel2020self}
Alwassel, H., Korbar, B., Mahajan, D., Torresani, L., Ghanem, B., Tran, D.:
  Self-supervised learning by cross-modal audio-video clustering. In: NeurIPS
  (2020)

\bibitem{asano2020labelling}
Asano, Y.M., Patrick, M., Rupprecht, C., Vedaldi, A.: Labelling unlabelled
  videos from scratch with multi-modal self-supervision. In: NeurIPS (2020)

\bibitem{asano2021pass}
Asano, Y.M., Rupprecht, C., Zisserman, A., Vedaldi, A.: Pass: An imagenet
  replacement for self-supervised pretraining without humans. In: Thirty-fifth
  Conference on Neural Information Processing Systems Datasets and Benchmarks
  Track (Round 1) (2021)

\bibitem{bain2021frozen}
Bain, M., Nagrani, A., Varol, G., Zisserman, A.: Frozen in time: A joint video
  and image encoder for end-to-end retrieval. arXiv preprint arXiv:2104.00650
  (2021)

\bibitem{bertasius2021space}
Bertasius, G., Wang, H., Torresani, L.: Is space-time attention all you need
  for video understanding? arXiv preprint arXiv:2102.05095  (2021)

\bibitem{carreira2019short}
Carreira, J., Noland, E., Hillier, C., Zisserman, A.: A short note on the
  kinetics-700 human action dataset. arXiv preprint arXiv:1907.06987  (2019)

\bibitem{carreira2017quo}
Carreira, J., Zisserman, A.: Quo vadis, action recognition? a new model and the
  kinetics dataset. In: proceedings of the IEEE Conference on Computer Vision
  and Pattern Recognition. pp. 6299--6308 (2017)

\bibitem{chen2020vggsound}
Chen, H., Xie, W., Vedaldi, A., Zisserman, A.: Vggsound: A large-scale
  audio-visual dataset. In: ICASSP 2020-2020 IEEE International Conference on
  Acoustics, Speech and Signal Processing (ICASSP). pp. 721--725. IEEE (2020)

\bibitem{chen2020generative}
Chen, M., Radford, A., Child, R., Wu, J., Jun, H.: Generative pretraining from
  pixels. In: ICML (2020)

\bibitem{chen2019uniter}
Chen, Y.C., Li, L., Yu, L., Kholy, A.E., Ahmed, F., Gan, Z., Cheng, Y., Liu,
  J.: Uniter: Learning universal image-text representations. arXiv preprint
  arXiv:1909.11740  (2019)

\bibitem{clark2020electra}
Clark, K., Luong, M.T., Le, Q.V., Manning, C.D.: {ELECTRA}: Pre-training text
  encoders as discriminators rather than generators. In: ICLR (2020)

\bibitem{desai2020virtex}
Desai, K., Johnson, J.: Virtex: Learning visual representations from textual
  annotations. arXiv preprint arXiv:2006.06666  (2020)

\bibitem{desai2021redcaps}
Desai, K., Kaul, G., Aysola, Z., Johnson, J.: Redcaps: Web-curated image-text
  data created by the people, for the people. arXiv preprint arXiv:2111.11431
  (2021)

\bibitem{devlin2019bert}
Devlin, J., Chang, M.W., Lee, K., Toutanova, K.: Bert: Pre-training of deep
  bidirectional transformers for language understanding. In: ACL (2019)

\bibitem{fang2021clip2video}
Fang, H., Xiong, P., Xu, L., Chen, Y.: Clip2video: Mastering video-text
  retrieval via image clip. arXiv preprint arXiv:2106.11097  (2021)

\bibitem{gebru2021datasheets}
Gebru, T., Morgenstern, J., Vecchione, B., Vaughan, J.W., Wallach, H., Iii,
  H.D., Crawford, K.: Datasheets for datasets. Communications of the ACM
  \textbf{64}(12),  86--92 (2021)

\bibitem{halevy2020preserving}
Halevy, A., Ferrer, C.C., Ma, H., Ozertem, U., Pantel, P., Saeidi, M.,
  Silvestri, F., Stoyanov, V.: Preserving integrity in online social networks.
  arXiv preprint arXiv:2009.10311  (2020)

\bibitem{Detoxify}
Hanu, L., {Unitary team}: Detoxify. Github.
  https://github.com/unitaryai/detoxify (2020)

\bibitem{he2016deep}
He, K., Zhang, X., Ren, S., Sun, J.: Deep residual learning for image
  recognition. In: Proceedings of the IEEE conference on computer vision and
  pattern recognition. pp. 770--778 (2016)

\bibitem{hendricks18emnlp}
Hendricks, L.A., Wang, O., Shechtman, E., Sivic, J., Darrell, T., Russell, B.:
  Localizing moments in video with temporal language. In: Empirical Methods in
  Natural Language Processing (EMNLP) (2018)

\bibitem{huang2021multilingual}
Huang, P.Y., Patrick, M., Hu, J., Neubig, G., Metze, F., Hauptmann, A.:
  Multilingual multimodal pre-training for zero-shot cross-lingual transfer of
  vision-language models. In: Meeting of the North American Chapter of the
  Association for Computational Linguistics (NAACL) (June 2021)

\bibitem{JDH17}
Johnson, J., Douze, M., J{\'e}gou, H.: Billion-scale similarity search with
  gpus. arXiv preprint arXiv:1702.08734  (2017)

\bibitem{kinetics}
Kay, W., Carreira, J., Simonyan, K., Zhang, B., Hillier, C., Vijayanarasimhan,
  S., Viola, F., Green, T., Back, T., Natsev, P., et~al.: The kinetics human
  action video dataset. arXiv preprint arXiv:1705.06950  (2017)

\bibitem{kingma2014adam}
Kingma, D.P., Ba, J.: Adam: A method for stochastic optimization. In: ICLR
  (2015)

\bibitem{krishna2017dense}
Krishna, R., Hata, K., Ren, F., Fei-Fei, L., Carlos~Niebles, J.:
  Dense-captioning events in videos. In: CVPR (2017)

\bibitem{kuznetsova2020open}
Kuznetsova, A., Rom, H., Alldrin, N., Uijlings, J., Krasin, I., Pont-Tuset, J.,
  Kamali, S., Popov, S., Malloci, M., Kolesnikov, A., et~al.: The open images
  dataset v4. International Journal of Computer Vision  \textbf{128}(7),
  1956--1981 (2020)

\bibitem{lei2021understanding}
Lei, C., Luo, S., Liu, Y., He, W., Wang, J., Wang, G., Tang, H., Miao, C., Li,
  H.: Understanding chinese video and language via contrastive multimodal
  pre-training. In: Proceedings of the 29th ACM International Conference on
  Multimedia. pp. 2567--2576 (2021)

\bibitem{lei2021less}
Lei, J., Li, L., Zhou, L., Gan, Z., Berg, T.L., Bansal, M., Liu, J.: Less is
  more: {ClipBERT} for video-and-language learning via sparse sampling. CVPR
  (2021)

\bibitem{lewis2020pretraining}
Lewis, M., Ghazvininejad, M., Ghosh, G., Aghajanyan, A., Wang, S., Zettlemoyer,
  L.: Pre-training via paraphrasing. arXiv preprint arXiv:2006.15020  (2020)

\bibitem{lewis2019bart}
Lewis, M., Liu, Y., Goyal, N., Ghazvininejad, M., Mohamed, A., Levy, O.,
  Stoyanov, V., Zettlemoyer, L.: Bart: Denoising sequence-to-sequence
  pre-training for natural language generation, translation, and comprehension.
  In: ACL (2020)

\bibitem{li2020unicoder}
Li, G., Duan, N., Fang, Y., Gong, M., Jiang, D., Zhou, M.: Unicoder-vl: A
  universal encoder for vision and language by cross-modal pre-training. In:
  AAAI (2020)

\bibitem{li2022blip}
Li, J., Li, D., Xiong, C., Hoi, S.: Blip: Bootstrapping language-image
  pre-training for unified vision-language understanding and generation (2022)

\bibitem{li2020hero}
Li, L., Chen, Y.C., Cheng, Y., Gan, Z., Yu, L., Liu, J.: Hero: Hierarchical
  encoder for video+ language omni-representation pre-training. EMNLP  (2020)

\bibitem{li2019visualbert}
Li, L.H., Yatskar, M., Yin, D., Hsieh, C.J., Chang, K.W.: Visualbert: A simple
  and performant baseline for vision and language. arXiv preprint
  arXiv:1908.03557  (2019)

\bibitem{lin2014microsoft}
Lin, T.Y., Maire, M., Belongie, S., Hays, J., Perona, P., Ramanan, D.,
  Doll{\'a}r, P., Zitnick, C.L.: Microsoft coco: Common objects in context. In:
  European conference on computer vision. pp. 740--755. Springer (2014)

\bibitem{lu2019vilbert}
Lu, J., Batra, D., Parikh, D., Lee, S.: Vilbert: Pretraining task-agnostic
  visiolinguistic representations for vision-and-language tasks. In: NeurIps
  (2019)

\bibitem{luo2020univl}
Luo, H., Ji, L., Shi, B., Huang, H., Duan, N., Li, T., Li, J., Bharti, T.,
  Zhou, M.: {UniVL}: A unified video and language pre-training model for
  multimodal understanding and generation. arXiv preprint arXiv:2002.06353
  (2020)

\bibitem{luo2021clip4clip}
Luo, H., Ji, L., Zhong, M., Chen, Y., Lei, W., Duan, N., Li, T.: Clip4clip: An
  empirical study of clip for end to end video clip retrieval. arXiv preprint
  arXiv:2104.08860  (2021)

\bibitem{livebot}
Ma, S., Cui, L., Dai, D., Wei, F., Sun, X.: Livebot: Generating live video
  comments based on visual and textual contexts. In: {AAAI} 2019 (2019)

\bibitem{miech2019howto100m}
Miech, A., Zhukov, D., Alayrac, J.B., Tapaswi, M., Laptev, I., Sivic, J.:
  Howto100m: Learning a text-video embedding by watching hundred million
  narrated video clips. In: ICCV (2019)

\bibitem{morgado2020avid}
Morgado, P., Vasconcelos, N., Misra, I.: Audio-visual instance discrimination
  with cross-modal agreement. arXiv preprint arXiv:2004.12943  (2020)

\bibitem{mu2021slip}
Mu, N., Kirillov, A., Wagner, D., Xie, S.: Slip: Self-supervision meets
  language-image pre-training. arXiv preprint arXiv:2112.12750  (2021)

\bibitem{patrick2020multimodal}
Patrick, M., Asano, Y.M., Kuznetsova, P., Fong, R., Henriques, J.F., Zweig, G.,
  Vedaldi, A.: Multi-modal self-supervision from generalized data
  transformations (2021)

\bibitem{patrick2020support}
Patrick, M., Huang, P.Y., Asano, Y., Metze, F., Hauptmann, A., Henriques, J.,
  Vedaldi, A.: Support-set bottlenecks for video-text representation learning.
  arXiv preprint arXiv:2010.02824  (2020)

\bibitem{radford2021learning}
Radford, A., Kim, J.W., Hallacy, C., Ramesh, A., Goh, G., Agarwal, S., Sastry,
  G., Askell, A., Mishkin, P., Clark, J., et~al.: Learning transferable visual
  models from natural language supervision. arXiv preprint arXiv:2103.00020
  (2021)

\bibitem{radford2019language}
Radford, A., Wu, J., Child, R., Luan, D., Amodei, D., Sutskever, I.: Language
  models are unsupervised multitask learners. OpenAI Blog  \textbf{1}(8), ~9
  (2019)

\bibitem{raffel2019exploring}
Raffel, C., Shazeer, N., Roberts, A., Lee, K., Narang, S., Matena, M., Zhou,
  Y., Li, W., Liu, P.J.: Exploring the limits of transfer learning with a
  unified text-to-text transformer. arXiv preprint arXiv:1910.10683  (2019)

\bibitem{rebuffi2017learning}
Rebuffi, S.A., Bilen, H., Vedaldi, A.: Learning multiple visual domains with
  residual adapters. In: NeurIPS (2017)

\bibitem{lsmdc}
Rohrbach, A., Torabi, A., Rohrbach, M., Tandon, N., Pal, C., Larochelle, H.,
  Courville, A., Schiele, B.: Movie description. International Journal of
  Computer Vision  (2017)

\bibitem{ruan2022survey}
Ruan, L., Jin, Q.: Survey: Transformer based video-language pre-training. AI
  Open  (2022)

\bibitem{sariyildiz2020learning}
Sariyildiz, M.B., Perez, J., Larlus, D.: Learning visual representations with
  caption annotations. In: ECCV. pp. 153--170. Springer (2020)

\bibitem{sharma2018conceptual}
Sharma, P., Ding, N., Goodman, S., Soricut, R.: Conceptual captions: A cleaned,
  hypernymed, image alt-text dataset for automatic image captioning. In:
  Proceedings of the 56th Annual Meeting of the Association for Computational
  Linguistics (Volume 1: Long Papers). pp. 2556--2565 (2018)

\bibitem{su2019vlbert}
Su, W., Zhu, X., Cao, Y., Li, B., Lu, L., Wei, F., Dai, J.: {VL-BERT}:
  Pre-training of generic visual-linguistic representations. arXiv preprint
  arXiv:1908.08530  (2019)

\bibitem{sun2019learning}
Sun, C., Baradel, F., Murphy, K., Schmid, C.: Learning video representations
  using contrastive bidirectional transformer. arXiv preprint arXiv:1906.05743
  (2019)

\bibitem{Sun_2019}
Sun, C., Myers, A., Vondrick, C., Murphy, K., Schmid, C.: Videobert: A joint
  model for video and language representation learning. In: ICCV (2019)

\bibitem{Tan_2019}
Tan, H., Bansal, M.: Lxmert: Learning cross-modality encoder representations
  from transformers. In: EMNLP (2019)

\bibitem{thomee2016yfcc100m}
Thomee, B., Shamma, D.A., Friedland, G., Elizalde, B., Ni, K., Poland, D.,
  Borth, D., Li, L.J.: Yfcc100m: The new data in multimedia research.
  Communications of the ACM  \textbf{59}(2),  64--73 (2016)

\bibitem{vaswani2017attention}
Vaswani, A., Shazeer, N., Parmar, N., Uszkoreit, J., Jones, L., Gomez, A.N.,
  Kaiser, L., Polosukhin, I.: Attention is all you need. In: NeurIPS (2017)

\bibitem{msvd}
Venugopalan, S., Rohrbach, M., Donahue, J., Mooney, R., Darrell, T., Saenko,
  K.: Sequence to sequence -- video to text. In: Proceedings of the IEEE
  International Conference on Computer Vision (ICCV) (2015)

\bibitem{wu2020response}
Wu, H., Jones, G.J., Pitie, F.: Response to livebot: Generating live video
  comments based on visual and textual contexts. arXiv preprint
  arXiv:2006.03022  (2020)

\bibitem{vtt}
Xu, J., Mei, T., Yao, T., Rui, Y.: {MSR-VTT:} {A} large video description
  dataset for bridging video and language. In: CVPR (2016)

\bibitem{yao2022filip}
Yao, L., Huang, R., Hou, L., Lu, G., Niu, M., Xu, H., Liang, X., Li, Z., Jiang,
  X., Xu, C.: {FILIP}: Fine-grained interactive language-image pre-training.
  In: International Conference on Learning Representations (2022)

\bibitem{zhu2020actbert}
Zhu, L., Yang, Y.: Actbert: Learning global-local video-text representations.
  In: CVPR (2020)

\end{thebibliography}
\clearpage

\section{Appendix}

\subsection{Qualitative Examples}

\begin{figure}[h]
    \centering
    \includegraphics[scale=0.57]{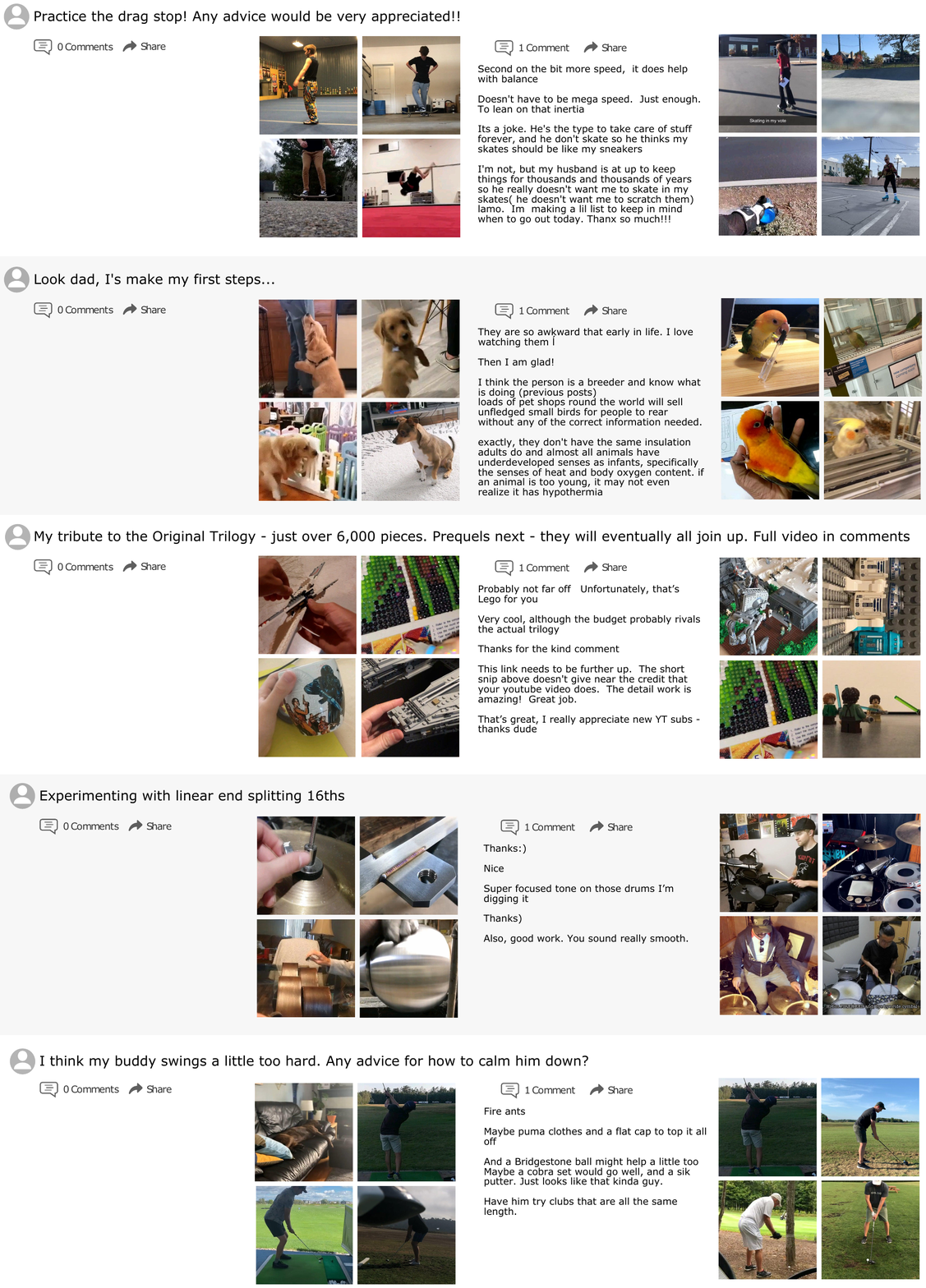}
    \caption{Examples of retrieved video thumbnails when adapting the text branch.}
    \label{fig:adapting_title_examples}
\end{figure}

In \cref{fig:adapting_title_examples} we adapt the text branch, similar to Fig.~1 of the main paper.
The example in the second row of \cref{fig:adapting_title_examples} shows how our Context Adapter Module can leverage the comments to learn that the content is indeed about parrots, as opposed to dogs.
The fourth row shows that without comments, the title alone can be extremely ambiguous while comments can again guide the model to retrieve relevant videos of drumming.

\begin{figure}[h]
    \centering
    \includegraphics[scale=0.57]{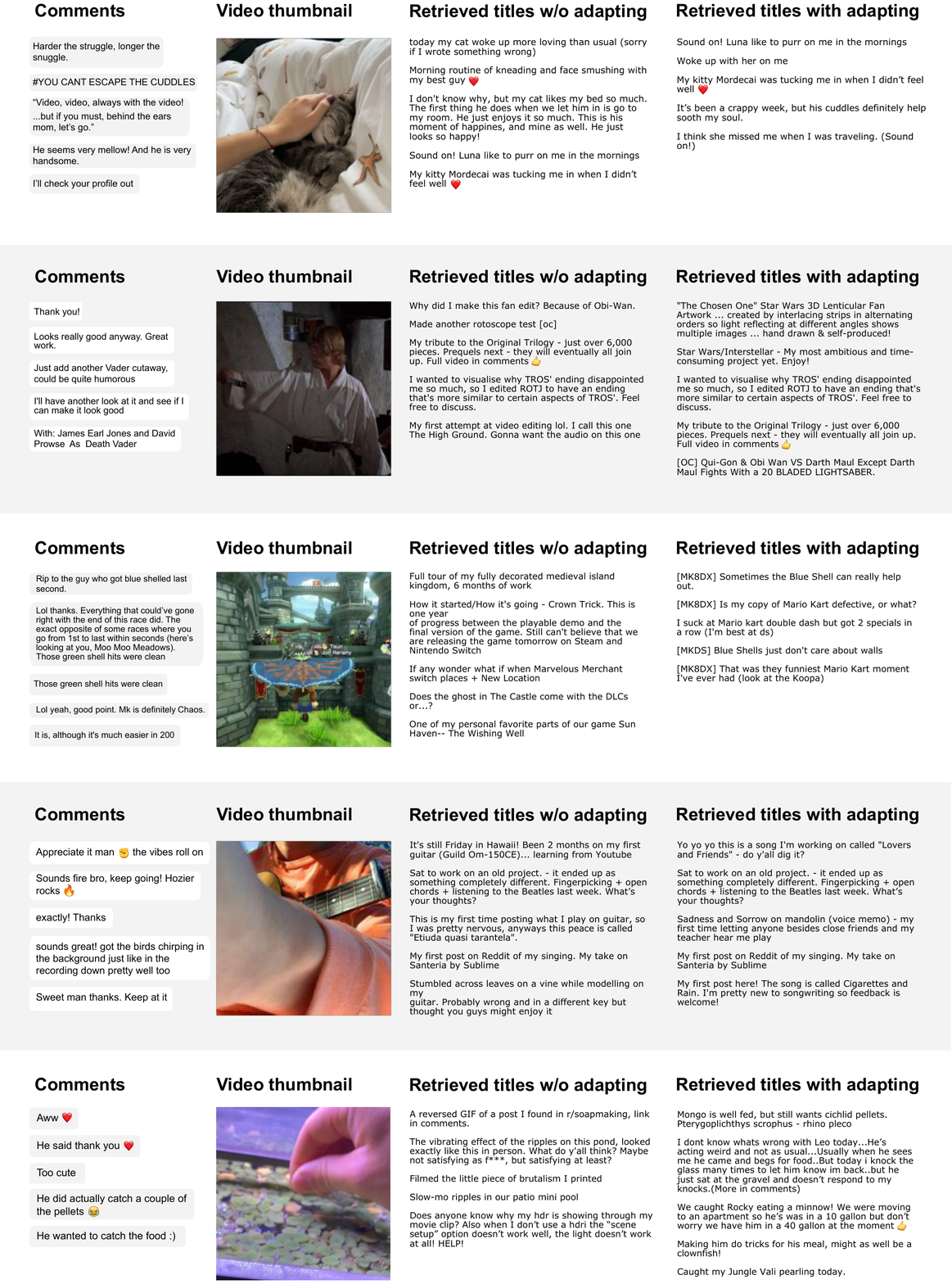}
    \caption{Examples of retrieved titles when adapting the visual branch.}
    \label{fig:adapting_image_examples}
\end{figure}

We provide examples where the video branch has been adapted in \cref{fig:adapting_image_examples}. In most cases, the retrieved titles are broadly related to the video thumbnails. However, when provided with the comments, the retrieved titles become more specific to the videos. For example, in the example from the second row of a screenshot from Mario Kart, the retrieved titles are generally about games e.g. The Castle or Sun Haven, whereas when adapting the video with the comments, the model retrieves titles specifically about Mario Kart. Similarly, in the example from the last row, the model seems to get confused about the content of the video when deprived of the comments, which provide the necessary context about feeding a fish.

\begin{figure}[h]
    \centering
    \includegraphics[scale=0.38]{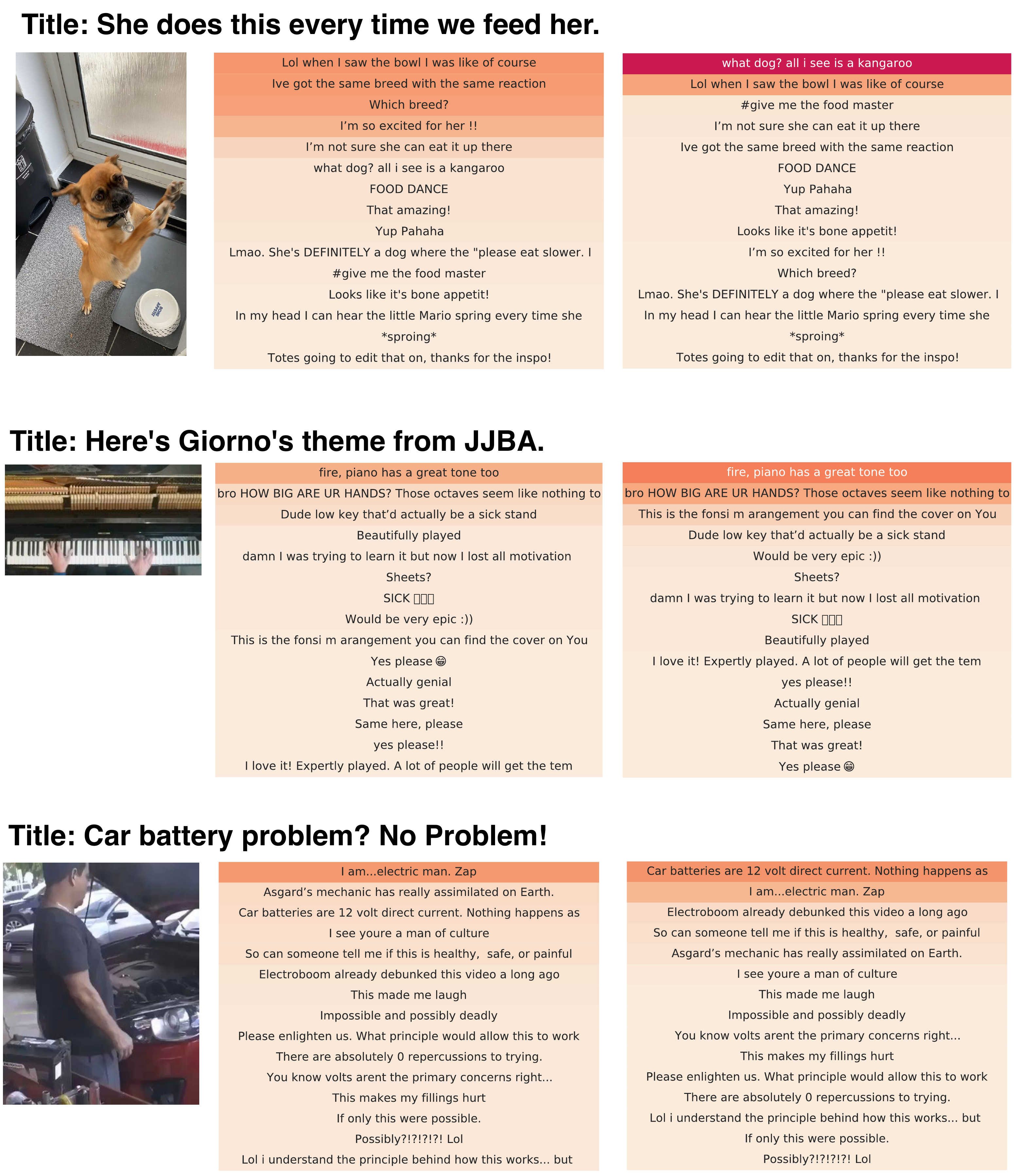}
    \caption{\textbf{Visualising comment saliency.} We show the title and thumbnail for three videos, and
    show the ranked saliency of comments when adapting using the Text branch (left) and Image branch (right). Comments mentioning
    topics relevant to the title or image are ranked highly, while irrelevant  comments are lower.}
    \label{fig:comment_saliency}
\end{figure}

Finally, in \cref{fig:comment_saliency}, we show the saliency of comments with regards to a given video and title.
For this, we use the approach of masking out each comment in turn, allowing us to visualise the effect of each individual
comment on the network output. We compare the output descriptor when including all comments to the descriptors with a
comment masked, using the inner product as a score of similarity, and present the comments sorted from lowest to high,
the expectation being that an uninformative comment will not cause a large shift in the descriptor (so will still have high similarity
when excluded) whereas a salient comment will cause a larger shift (and so a lower similarity when excluded). We show results for adapting
both the text branch (left) and visual branch (right), and observe that, as expected, uninformative comments such as
``\textit{That was great!}'' and ``\textit{Possibly?!?!?!?! Lol}'' cause little change to the descriptor, whereas comments related to objects
in the video cause a larger shift. This demonstrates that the method is able to pick out and filter the relevant information.

\begin{table}[htb]
    \footnotesize
    \centering
    \setlength{\tabcolsep}{6pt}
    \captionof{table}{\textbf{Video results - adapting the text branch} We try adapting the text branch rather than the video branch for the video experiment. In this setting, the addition of comments seems to transfer less well to other datasets. Showing Recall@10}
    \begin{tabular}[t]{@{}llllllll@{}}
        \toprule
        & &  \multicolumn{2}{c}{\textbf{\dataset}} & \multicolumn{2}{c}{\textbf{KineticsComms}} & \multicolumn{2}{c}{\textbf{LiveBotEN}} \\
        \textbf{inference} & \textbf{\#frames}  & \textbf{VTR}  &  \textbf{TVR} & \textbf{VTR}  &  \textbf{TVR}  & \textbf{VTR}  &  \textbf{TVR} \\ 
        \midrule
        video           &    1  &   29.1  & 28.6  &  49.1   & 46.7   &  48.0 & 52.0    \\
        video+comments  &    1  &   33.2  &  33.5  &  47.8  &  45.6  &  49.0 & 52.0   \\
        \midrule
        video           &    8  &    28.6  &  27.8  &  57.5  &   55.3  & 68.0  &  71.0  \\
        video+comments  &    8  &    33.7  & 33.3   &  57.3  &  53.7   & 67.0  & 67.0 \\
        \bottomrule
    \end{tabular}
\label{tab:video_textbranch}
\end{table}

\subsection{Additional Results}
Similar to Tab.7 in the main paper, in \cref{tab:zeroshot} we evaluate zero-shot generalization of our video model on MSRVTT and MSVD (w/o comments) and compare to CLIP which has been shown to generalize very well \cite{radford2021learning}.

\begin{table}[htb]
    \footnotesize
    \centering
    \setlength{\tabcolsep}{6pt}
    \captionof{table}{\textbf{Zero-Shot Generalization.} Comparison of zero-shot generalization (without using comments). Results are TVR@10.}
    \begin{tabular}[t]{@{}lll@{}}
        \toprule
        &  \textbf{MSRVTT} & \textbf{MSVD} \\ 
        \midrule
        CLIP            &    60.7 & 65.27   \\
        Ours            &    63.8 & 76.93   \\
        \bottomrule
    \end{tabular}
\label{tab:zeroshot}
\end{table}

Additionally, we perform a baseline experiment by removing all visual information for retrieval in \cref{tab:textonly}. As expected, using the video with comments results in improved results over text-only retrieval.

\begin{table}[htb]
    \footnotesize
    \centering
    \setlength{\tabcolsep}{6pt}
    \captionof{table}{\textbf{Text-only baseline.} Comparing retrieval performance without any visual information.}
    \begin{tabular}[t]{@{}lll@{}}
        \toprule
        &  \textbf{R@1} & \textbf{R@10} \\ 
        \midrule
        Title from Comments &  20.3 & 41.3 \\
        Comments from Title &  20.0 & 42.2 \\ \midrule
        Title from Video & 28.2 & 51.2 \\
        Video from Title & 25.1 & 49.9 \\
        \bottomrule
    \end{tabular}
\label{tab:textonly}
\end{table}

\subsection{Dataset Statistics}

\begin{figure}[h]
    \centering
    \includegraphics[scale=0.1]{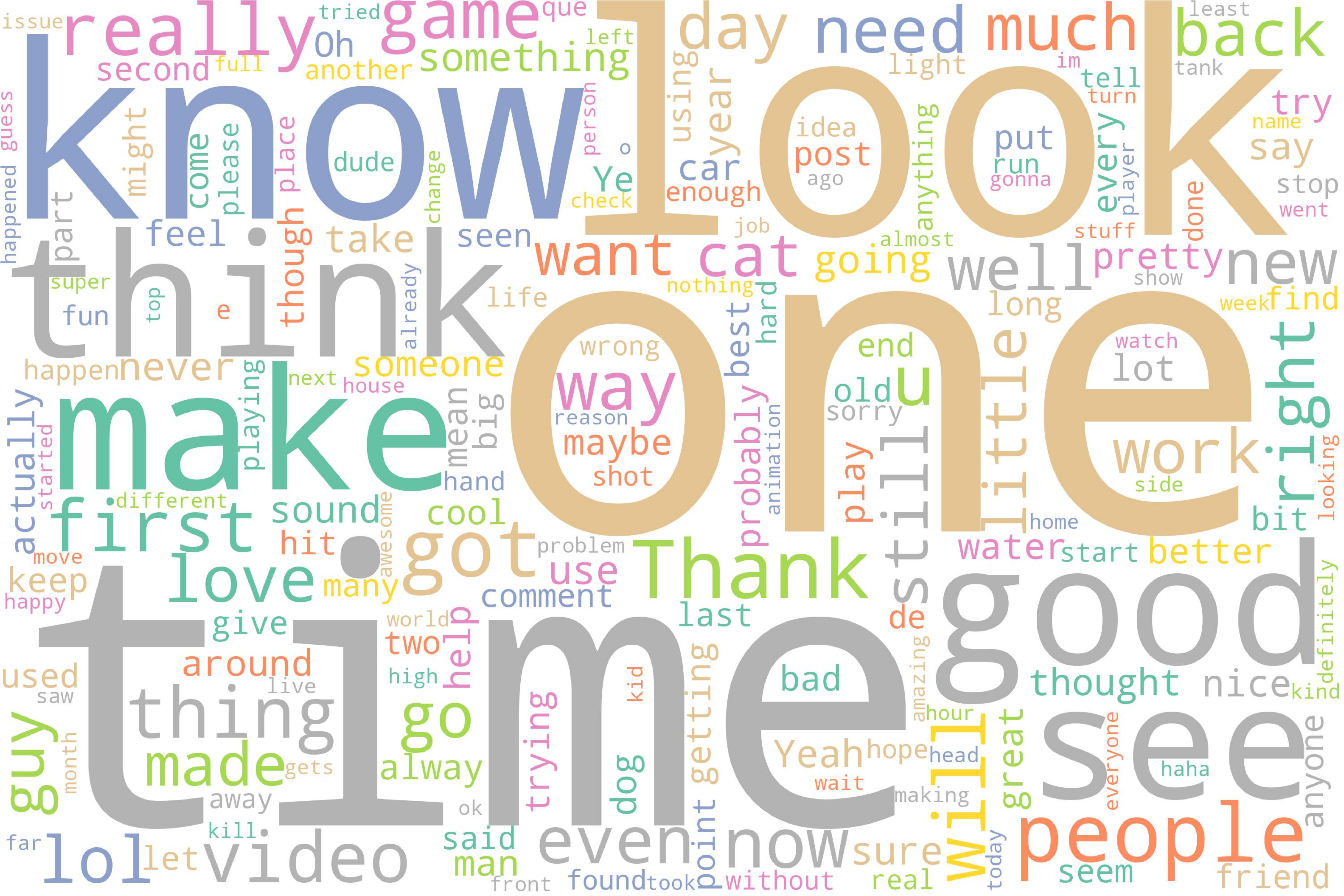}
    \caption{Word cloud of most common words in the captions.}
    \label{fig:wordcloud_caption}
\end{figure}

\begin{figure}[h]
    \centering
    \includegraphics[scale=0.1]{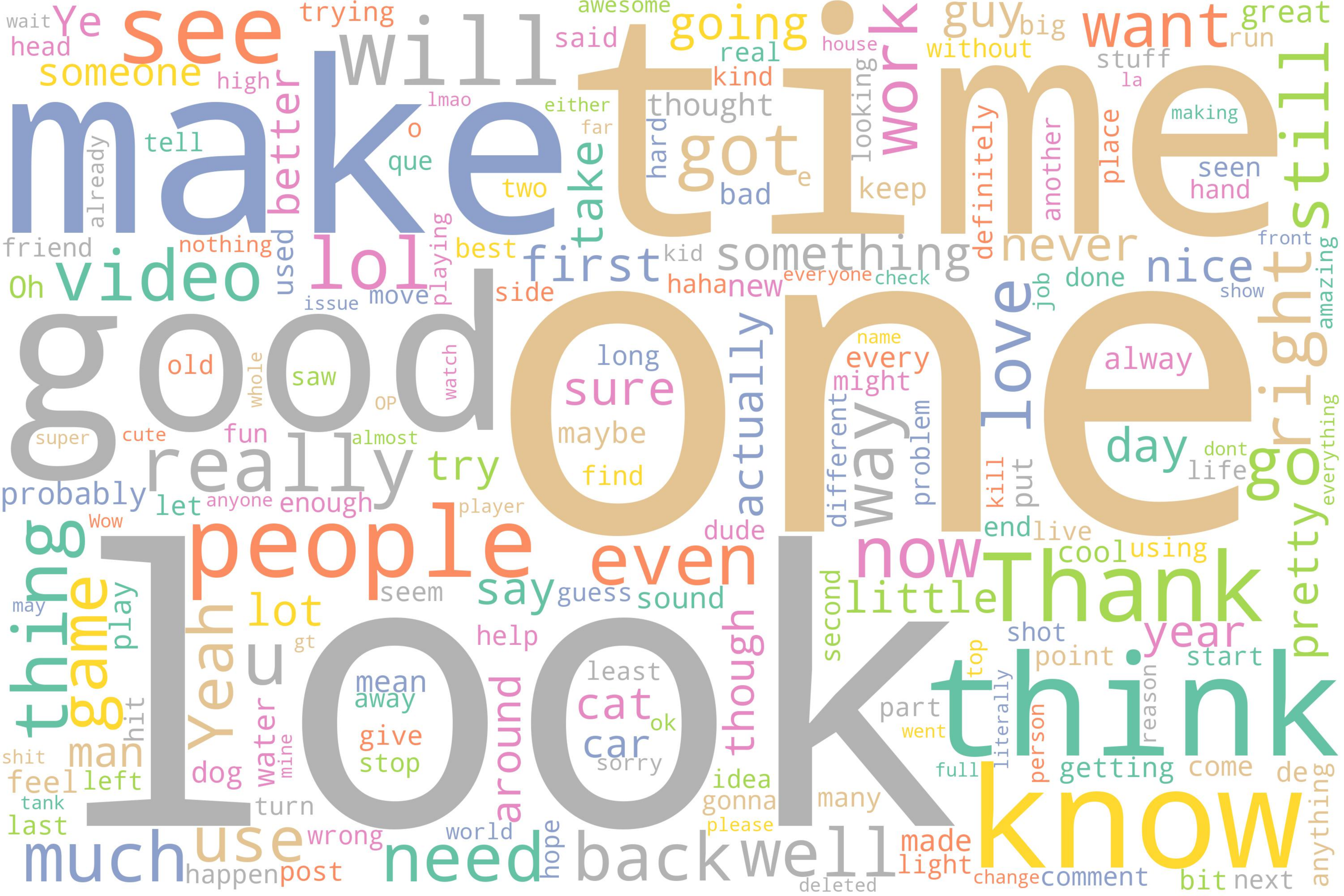}
    \caption{Word cloud of most common words in the comments.}
    \label{fig:wordcloud_comments}
\end{figure}

\begin{figure}[h]
    \centering
    \includegraphics[scale=0.6]{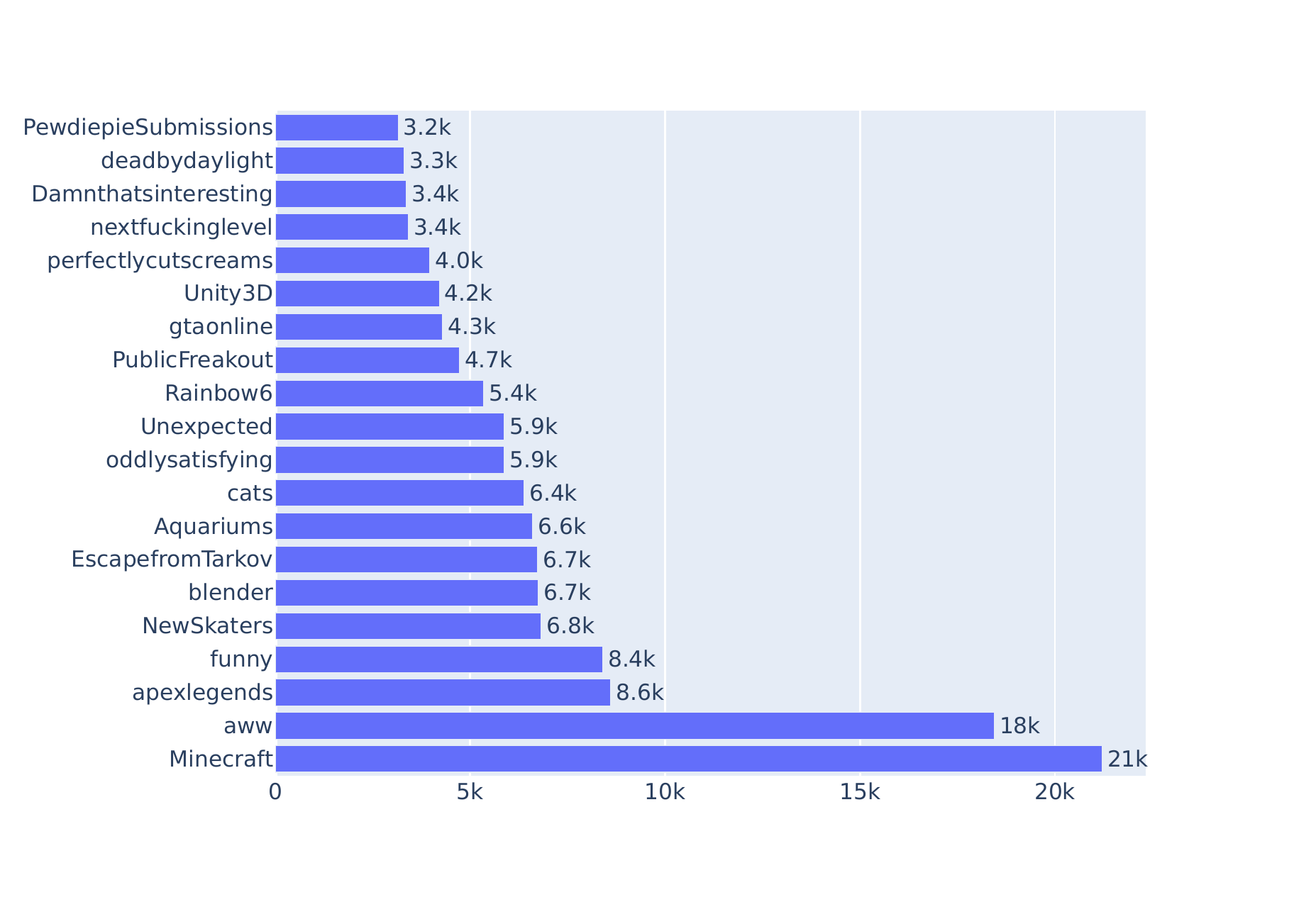}
    \caption{Top 20 subreddits according to the number of videos.}
    \label{fig:top_subreddit}
\end{figure}

\begin{figure}[h]
    \centering
    \includegraphics[scale=0.6]{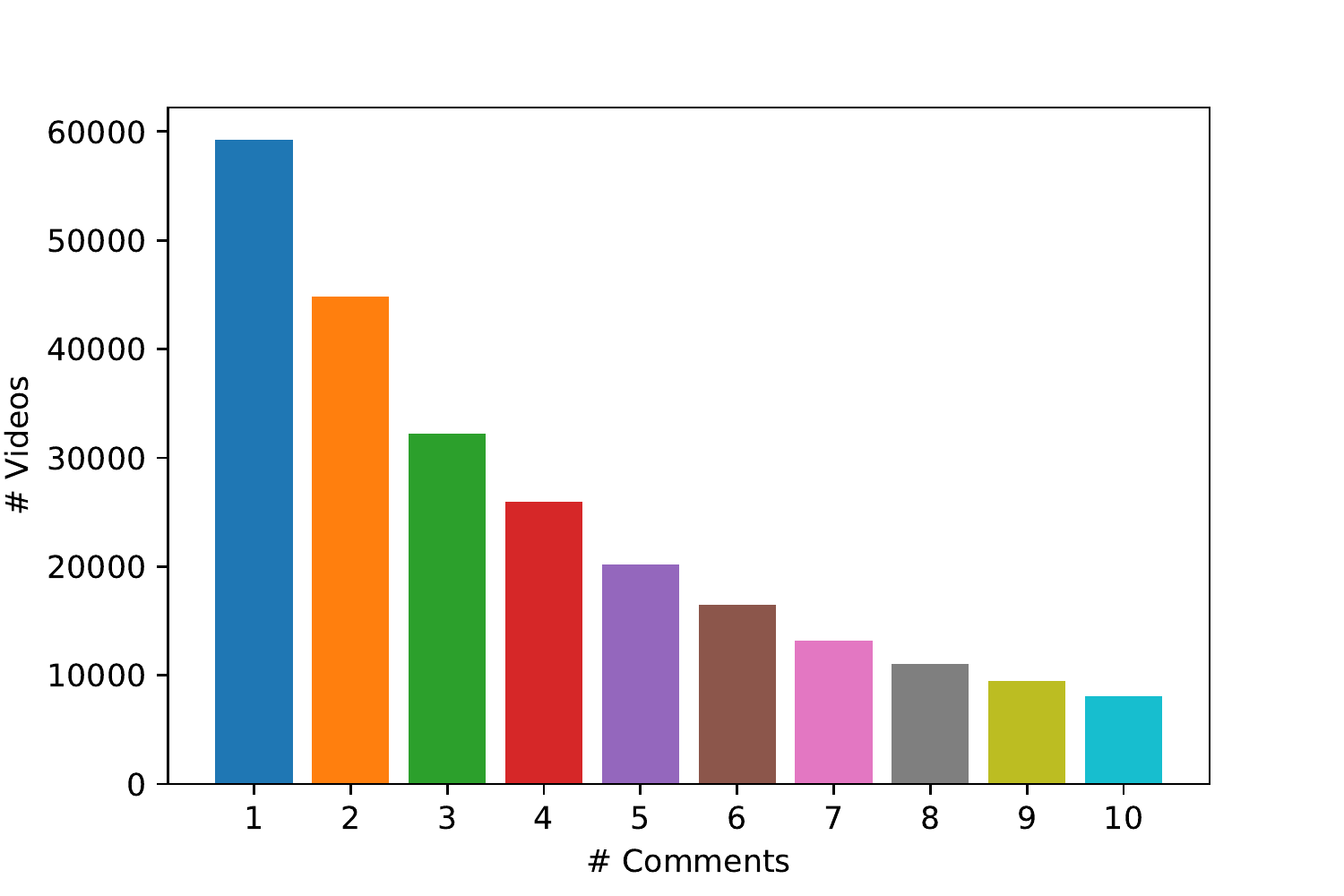}
    \caption{We show a histogram of comment statistics on VTC.}
    \label{fig:comment_stats_reddit}
\end{figure}

\begin{figure}[h]
    \centering
    \includegraphics[scale=0.6]{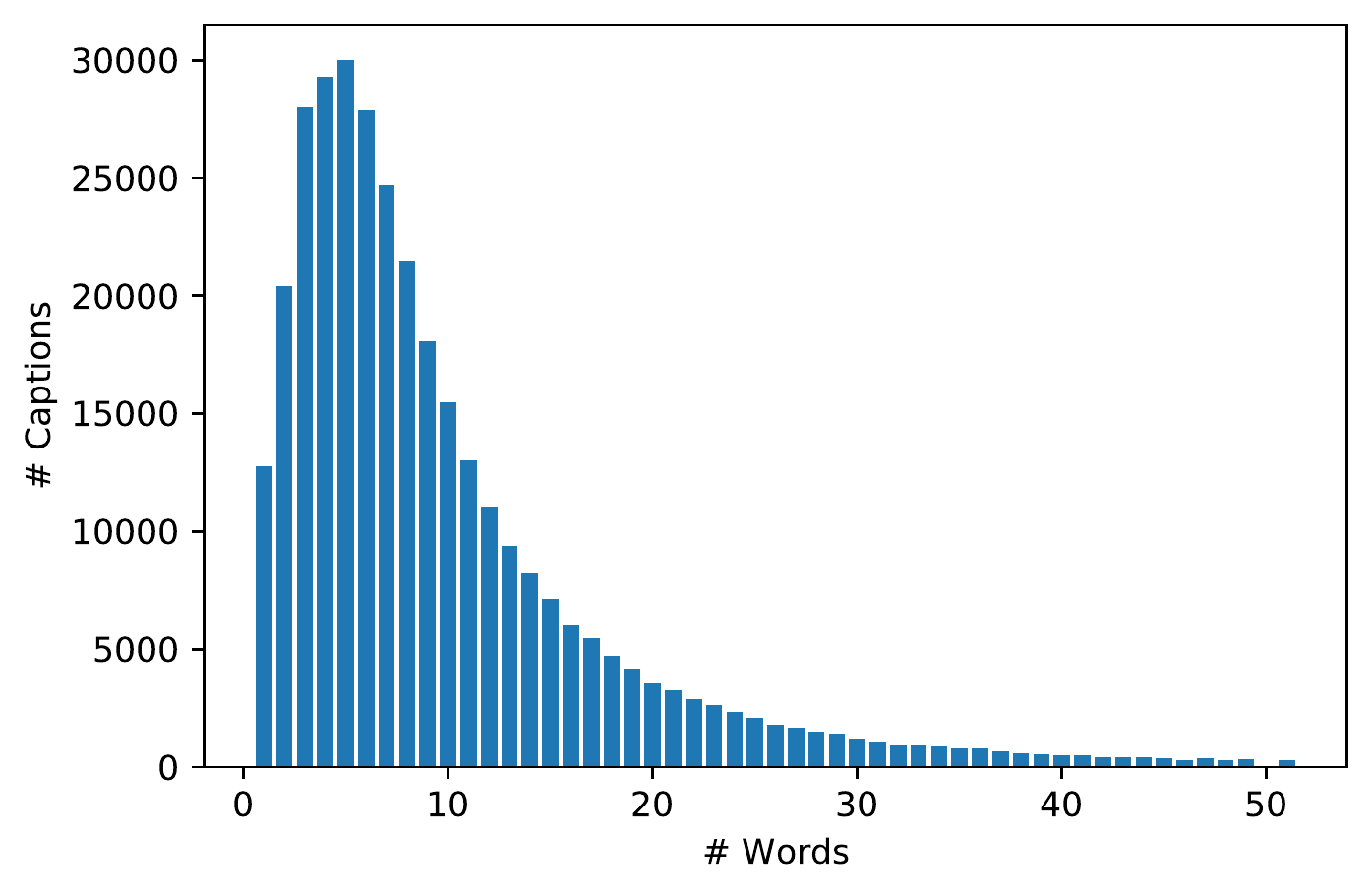}
    \caption{Caption length distribution.}
    \label{fig:word_freq_captions}
\end{figure}

\begin{figure}[h]
    \centering
    \includegraphics[scale=0.6]{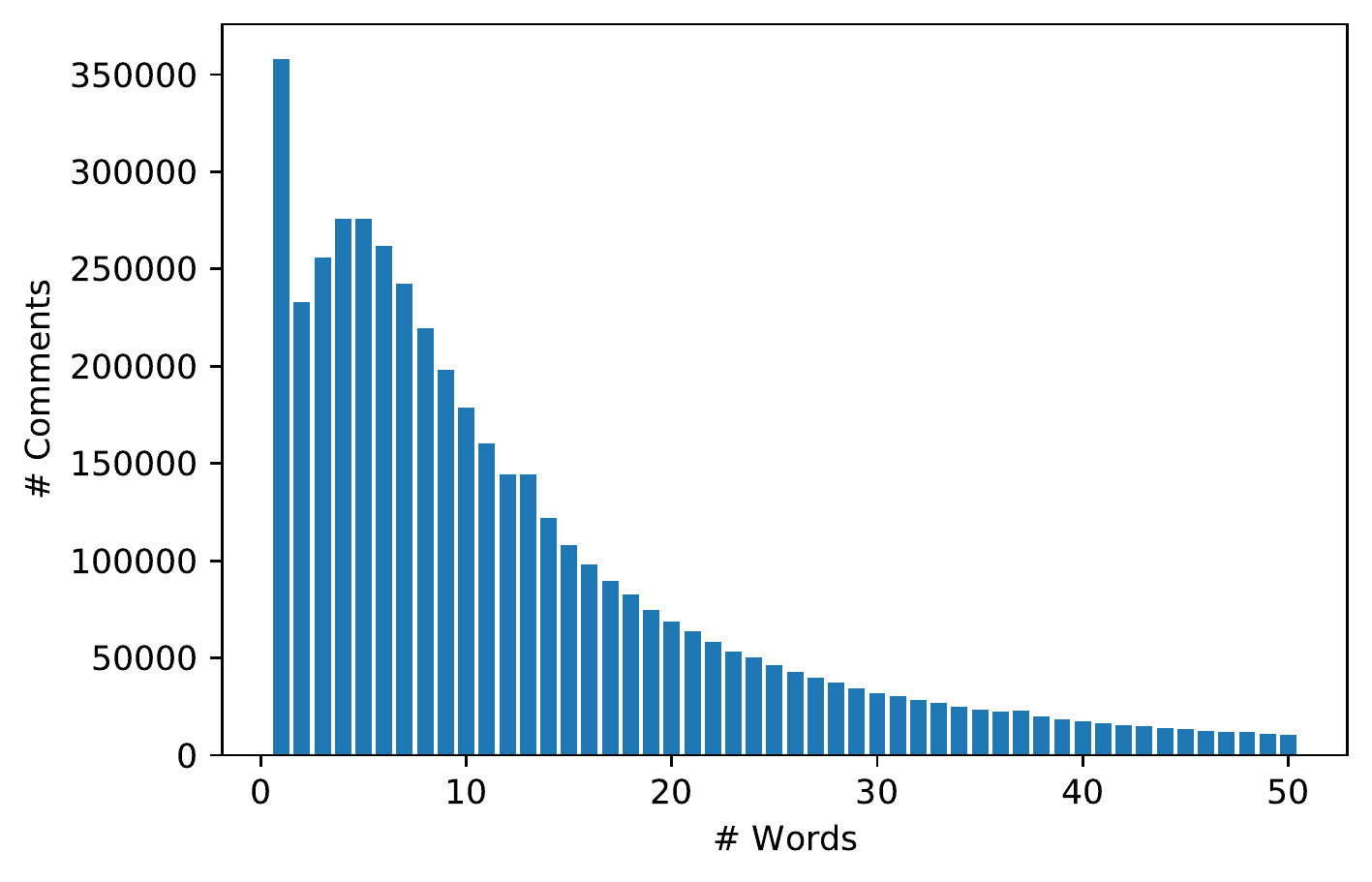}
    \caption{Comment length distribution.}
    \label{fig:word_freq_comments}
\end{figure}

In this section, we report some statistics of our dataset in order to give a sense of common topics and general distributions. 
We show word clouds of the most frequent words in the captions and  comments in figures \ref{fig:wordcloud_caption}-\ref{fig:wordcloud_comments}. In figure \ref{fig:top_subreddit}, we plot a histogram of most common subreddits based on the number of videos, with "Minecraft" having the largest proportion, followed closely by "aww". The distribution of the number of comments per post can be seen in \ref{fig:comment_stats_reddit}. In figures \ref{fig:word_freq_captions}-\ref{fig:word_freq_comments}, we show the distribution of captions and comment lengths, measured in number of words.

\subsection{Dataset Curation}

We use the GPU implementation of the FAISS similarity search toolkit \cite{JDH17} to efficiently deduplicate the dataset by indexing the video thumbnail embeddings obtained from a ResNet18. These indices are then used to discard video entries with a high similarity to other posts.

\subsection{Training Details}

The majority of experiments were conducted on a rented 4xA100-40GB GPU server costing approximately 170USD per day, over the course of three months.
Image models (using batch size 128) and video models without comments (using batch size 50) could train on a single 40GB GPU. For TimeSformer models the visual branch was processed on a separate GPU (when training with CAM and batch size 50) or pair of GPUs (for finetuning on video benchmarks with batch size 128). Pretraining the adapter on images takes approximately one hour per epoch. Training the full video model with CAM takes approximately 6 hours per epoch.
For the video experiments, we first train the CAM for 5 epochs with the backbone frozen, and then train the rest of the network for one epoch, with the backbone modified to have temporal attention.
We use the CAM with 5 comments, and adapt the visual branch of the model.

We use both photometric and temporal data augmentation. For photometric augmentation
we employ random crops ($0.5-1.0$ scale), random horizontal flipping, and colour jitter
(brightness, contrast, saturation, hue). For temporal augmentation, we first temporally
subsample the input frames (which are often 30fps) according to a random stride selected uniformly from 
$(4, 8, 16, 32)$ and then choose a random 8-frame segment uniformly. We normalise inputs using the
same preprocessing as Clip (ImageNet mean and standard deviation, $224\times224$ input size).

At evaluation time we use a temporal stride of 16 and split the video into 8-frame chunks,
taking the average of the descriptors of the chunks.

We randomly mask out comments with probability 0.5. We randomly skip adding the residual
from the adapter with probability 0.5, which ensures that unadapted descriptors are also
used in the loss and so the backbone network can still be used without the adapter.

All retrieval experiments are GPU accelerated using the FAISS\footnote{\url{https://github.com/facebookresearch/faiss}} library.

\subsection{Kinetics Comments} 

\begin{figure}[h]
    \centering
    \includegraphics[scale=0.7]{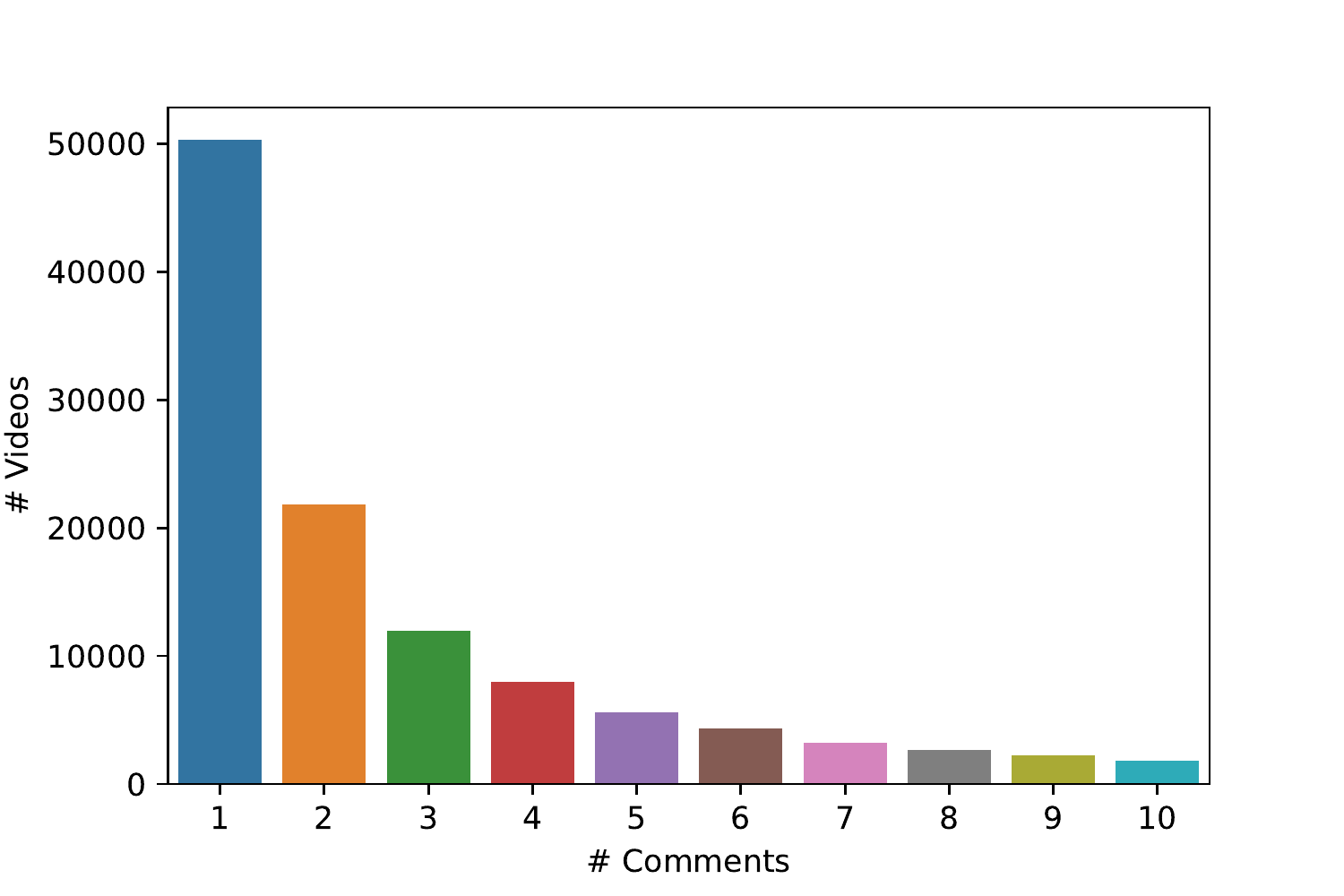}
    \caption{We show a histogram of comment statistics on KineticsComments.}
    \label{fig:kinstats}
\end{figure}

In this section we will describe the details for the additional comments we retrieve for the Kinectics-700 dataset. In \cref{tab:hist_kc} and \cref{fig:kinstats} we show the distribution of the number of comments in the dataset. We collect a maximum of 10 comments and exclude videos without comments, which leaves us with 111\,920 videos of the originally 650\,000 video clips. The majority of videos has one or two comments available.
\begin{table}
\footnotesize
\centering
\captionof{table}{Comments per video statistics for the KineticsComments dataset.} 
\begin{tabular}[h]{@{}lrrrrrrrrrr@{}}
\toprule
\textbf{\#comments}  & \textbf{1}  & \textbf{2}  & \textbf{3} & \textbf{4} & \textbf{5} & \textbf{6} & \textbf{7} & \textbf{8} & \textbf{9} & \textbf{10} \\ \midrule
\#videos & 50322&21847&11946&7960&5596&4311&3220&2671&2245&1852\\
\bottomrule
\end{tabular}
\label{tab:hist_kc}
\end{table}

\subsection{Additional failure cases}\label{sec:more_failure}
\begin{figure}[h]
    \centering
    \includegraphics[scale=0.49]{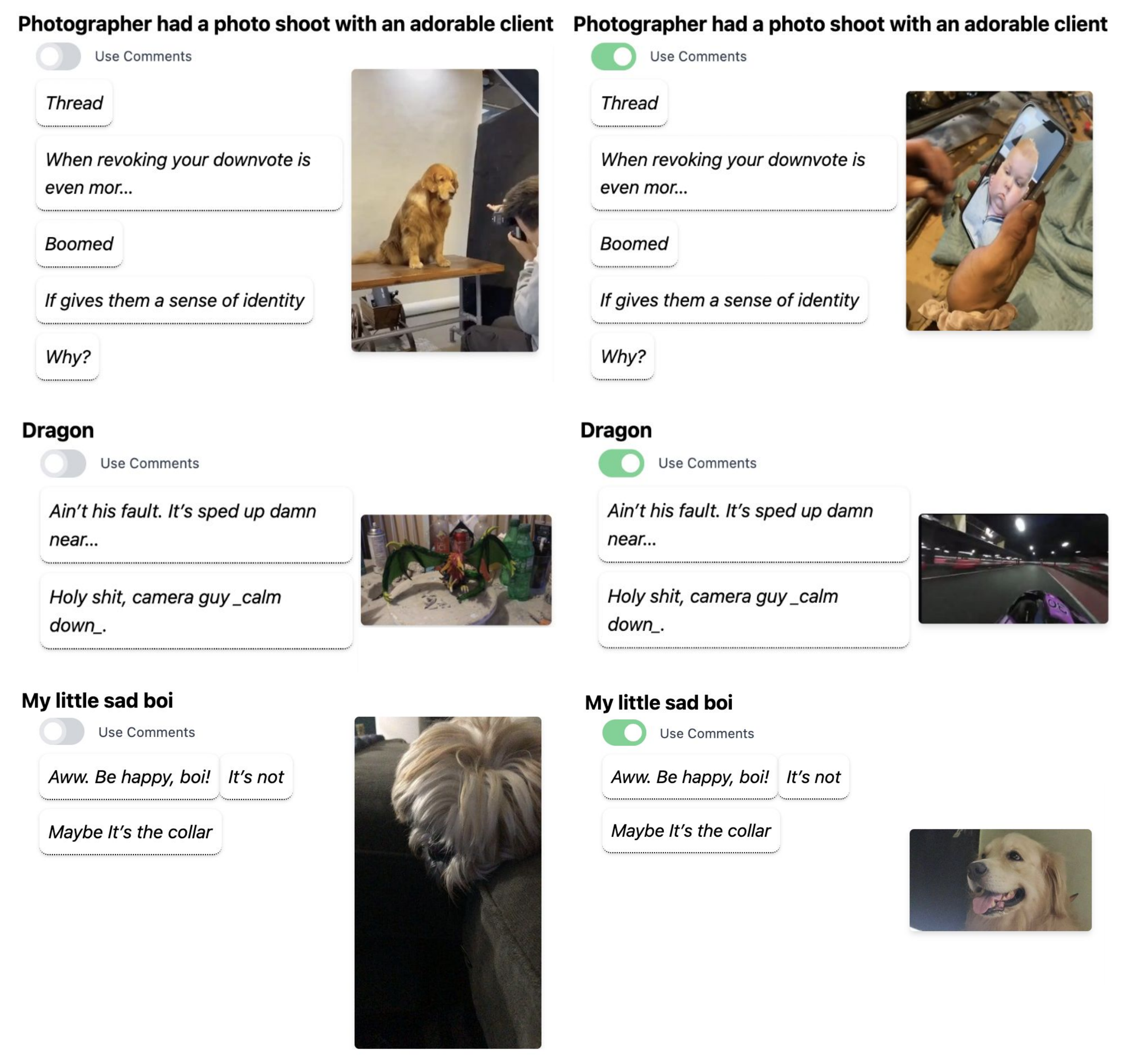} 
    \caption{ Examples of failure cases where using comments confounds the model and leads to a more mismatched retrieved thumbnail.}
    \label{fig:failure_cases}
\end{figure}

In \cref{fig:failure_cases} we show additional failure cases. We find that vary vague comments ``Why'' or generic expressions ``Ain't his fault'' can distract the model from the title.
In the last example, the model does not capture the concept of a sad dog due to the mention of ``happy'' in the comments. 

\color{black}

\section{Model Diagram}
\begin{figure}[h]
    \centering
    \includegraphics[scale=0.24]{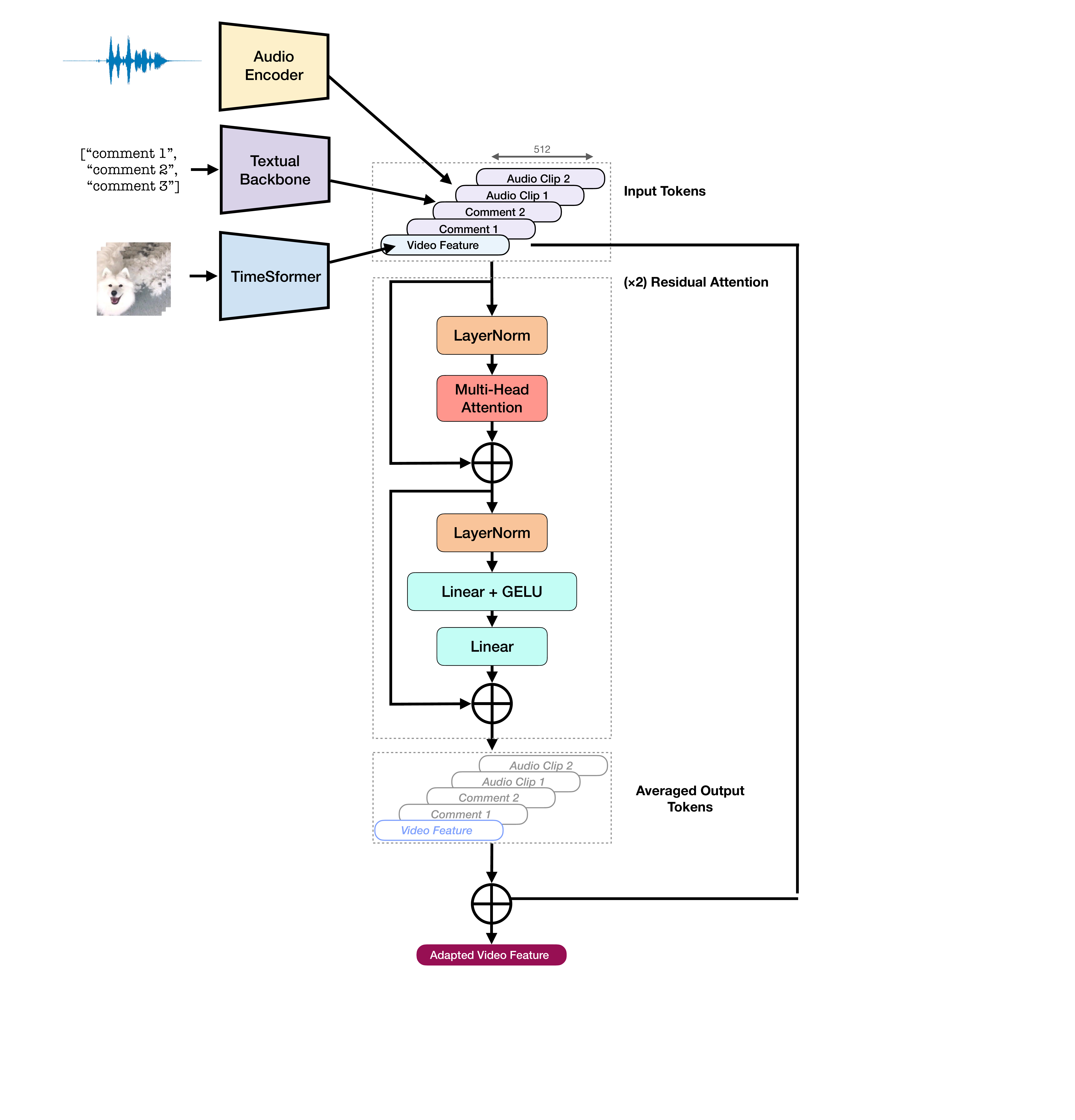}
    \caption{We show a diagram of the feature extraction and Context Adapter Module for the
    case of adapting the Video Feature. Multi-Head Self-Attention is performed on the input tokens
    (which are themselves any combination of video, audio or textual features) as part of a transformer architecture consisting
    of two Residual Attention blocks. Finally the output token corresponding to the Video Feature is
    passed through a final linear layer and added to the original feature in a residual fashion.}
    \label{fig:modeldiagram}
\end{figure}

\Cref{fig:modeldiagram} shows a diagram of the model.

\section{Datasheet for VTC dataset}

In this section, we answer the questions proposed by Gebru et al. in \cite{gebru2021datasheets}, which were introduced as a way of documenting new datasets.

\subsection{Motivation}

\paragraph{For what purpose was the dataset created?} 
The dataset was created strictly for research purposes. More specifically, this dataset addresses the research problem of using a weakly informative modality (user comments) in conjunction with other learning signals such as titles and videos for learning multi-modal representations.

\paragraph{Who created the dataset (e.g., which team, research group) and on behalf of which entity (e.g., company, institution, organization)?}
This dataset is created by VGG, a research group at the University of Oxford and Unitary AI, a company that's developing AI to automate content moderation.

\paragraph{Who funded the creation of the dataset?} 
The creation of dataset has not been funded directly. The individual researchers are funded by Amazon Machine Learning Awards (MLRA) and Innovate UK (project 71653) on behalf of UK Research and Innovation (UKRI).

\subsection{Composition}

\paragraph{What do the instances that comprise the dataset represent (e.g., documents, photos, people, countries)?}
The dataset is comprised of links to videos, titles, and comments. Each video-title pair corresponds to a post on reddit.com.
The dataset we share does not contain the data itself but hyperlinks to the data.

\paragraph{How many instances are there in total (of each type, if appropriate)?}
There are 339k video-title pairs with an average of 14 comments per video.

\paragraph{Does the dataset contain all possible instances or is it a sample (not necessarily random) of instances from a larger set?}

This dataset is a sample of a larger, unfiltered version of the original dataset that we have collected. From the initial version, we handpicked a list of "safe" subreddits and removed posts if: 1) they had the "NSFW" or "over\_18" tags; 2) the videos contained faces or the captions contained toxic or offensive text.

\paragraph{What data does each instance consist of?}
Each instance consists of:
- "reddit\_id"
- "post\_url"
- "comment\_ids"
- "subreddit"
- "video\_length"

\paragraph{Is there a label or target associated with each instance?}
No, there are no labels provided.

\paragraph{Is any information missing from individual instances?}
If a user decides to remove a post, the link to the post will become invalid and thus not accessible anymore.

\paragraph{Are relationships between individual instances made explicit (e.g., users’ movie ratings, social network links)?}
Instances that have the same subreddit are likely to share semantic meaning.

\paragraph{Are there recommended data splits (e.g., training, development/validation, testing)?}
We will release the data splits we have used in our experiments with our code.

\paragraph{Are there any errors, sources of noise, or redundancies in the dataset?}
Although we have tried to remove most bot-generated text, it is likely that some noise will still exist due to the nature of this data. Similarly, a small proportion of posts might still contain identical or highly similar videos post-deduplication.

\paragraph{Is the dataset self-contained, or does it link to or otherwise rely on external resources (e.g., websites, tweets, other datasets)?}
If it links to or relies on external resources, a) are there guarantees that they will exist, and remain constant, over time; b) are there official archival versions of the complete dataset (i.e., including the external resources as they existed at the time the dataset was created); c) are there any restrictions (e.g., licenses, fees) associated with any of the external resources that might apply to a future user? Please provide descriptions of all external resources and any restrictions associated with them, as well as links or other access points, as appropriate.

- In order to preserve user privacy, this dataset relies on links to reddit posts and comment ids. 
a) The links will no longer be valid if a user decides to delete their post.
b) It would be possible to find the metadata of each post, as well as the link to the media file, on the Reddit archive.
c) All links are accessible to everyone and are likely to remain so in the future.

\paragraph{Does the dataset contain data that might be considered confidential (e.g., data that is protected by legal privilege or by doctor-patient confidentiality, data that includes the content of individuals’ non-public communications)?}
No, all data shared links to public posts.

\paragraph{Does the dataset contain data that, if viewed directly, might be offensive, insulting, threatening, or might otherwise cause anxiety?}
The dataset is still likely to contain a small proportion of offensive data. Due to the size of the dataset, we were not able to verify each video and each comment manually. However, we have tried to minimize the number of unsafe posts by:
- manually filtering the subreddits included;
- using Reddit metadata such as the "NSFW" and "over\_18" tags to remove unsafe posts;
- using automatic machine learning models to remove posts containing faces and toxic text.

\paragraph{Does the dataset relate to people?}
The dataset relates to people in the sense that each post is created by a person. In order to minimise the content related to people, we used a public face detector model to remove most instances of videos containing faces.

\paragraph{Does the dataset identify any subpopulations (e.g., by age, gender)?}
The dataset does not explicitly identify any subpopulations. 
However, some titles, user comments or image contents may identify individuals as part of a subpopulation.

\paragraph{Is it possible to identify individuals (i.e., one or more natural persons), either directly or indirectly (i.e., in combination with other data) from the dataset?}
Yes. Our dataset contains links to posts where the Reddit username will be visible and some of them might have identifying information contained in their profile such as personal images or information. This information is, however, already publicly available on Reddit.

\paragraph{Does the dataset contain data that might be considered sensitive in any way (e.g., data that reveals racial or ethnic origins, sexual orientations, religious beliefs, political opinions or union memberships, or locations; financial or health data; biometric or genetic data; forms of government identification, such as social security numbers; criminal history)?}
While we believe this is highly unlikely (as we only use already public posts and comments) -- we cannot rule this out with absolute certainty. 
We will actively maintain this dataset after its release and ensure that if such information is included, that it is removed swiftly.

\subsection{Collection process}

\paragraph{How was the data associated with each instance acquired?}
The data was already available on Reddit.

\paragraph{What mechanisms or procedures were used to collect the data (e.g., hardware apparatus or sensor, manual human curation, software program, software API)?}
The dataset was collected via Reddit's own API (\url{https://www.reddit.com/wiki/api)}).

\paragraph{If the dataset is a sample from a larger set, what was the sampling strategy (e.g., deterministic, probabilistic with specific sampling probabilities)?}
NA

\paragraph{Who was involved in the data collection process (e.g., students, crowdworkers, contractors) and how were they compensated (e.g., how much were crowdworkers paid)?}
NA

\paragraph{Over what timeframe was the data collected? Does this timeframe match the creation timeframe of the data associated with the instances (e.g. recent crawl of old news articles)?} 
The dataset was collected between May 2020 and July 2021.

\paragraph{Were any ethical review processes conducted (e.g., by an institutional review board)?}
No.

\paragraph{Does the dataset relate to people?}
If not, you may skip the remainder of the questions in this section.
The dataset related to people in so far that the dataset creators are individual users of reddit and posts can contain people.

\paragraph{Did you collect the data from the individuals in question directly, or obtain it via third parties or other sources (e.g., websites)?}
The dataset was collected via Reddit's API. Thus, only public posts and data was downloaded.

\paragraph{Were the individuals in question notified about the data collection?}
NA.

\paragraph{Did the individuals in question consent to the collection and use of their data?}
NA.

\paragraph{If consent was obtained, were the consenting individuals provided with a mechanism to revoke their consent in the future or for certain uses?}
NA.

\paragraph{Has an analysis of the potential impact of the dataset and its use on data subjects (e.g., a data protection impact analysis) been conducted?}
NA.

\subsection{Preprocessing/cleaning/labeling}

\paragraph{Was any preprocessing/cleaning/labeling of the data done (e.g., discretization or bucketing, tokenization, part-of-speech tagging, SIFT feature extraction, removal of instances, processing of missing values)?}
The released dataset was preprocessed using an automated pipeline. 
This pipeline was taken from~\cite{asano2021pass} and was used to removed videos that contain human faces using a publicly available face classifier.

\paragraph{Was the “raw” data saved in addition to the preprocessed/cleaned/labeled data (e.g., to support unanticipated future uses)?}

Yes.

\subsection{Uses}

\paragraph{Has the dataset been used for any tasks already?}
This dataset has only been used for the experiments in this paper.
\paragraph{Is there a repository that links to any or all papers or systems that use the dataset?}
Google scholar will be able to track which papers have built upon this dataset/idea.

\paragraph{What (other) tasks could the dataset be used for?}
This dataset can be used for multi-modal representation learning or video-text retrieval.

\paragraph{Is there anything about the composition of the dataset or the way it was collected and preprocessed/cleaned/labeled that might impact future uses?}
Not that we are aware of.

\paragraph{Are there tasks for which the dataset should not be used?}
This dataset should not be used for tasks that might disclose the identity of the users or directly or indirectly harm them. 

\subsection{Distribution}

\paragraph{Will the dataset be distributed to third parties outside of the entity (e.g., company, institution, organization) on behalf of which the dataset was created?}
No.

\paragraph{How will the dataset will be distributed (e.g., tarball on website, API, GitHub)?}
The dataset will have a website and GitHub repository and be downloaded as a csv file containing links to the data points.

\paragraph{When will the dataset be distributed?}
The dataset will be published together with this paper.

\paragraph{Will the dataset be distributed under a copyright or other intellectual property (IP) license, and/or under applicable terms of use (ToU)?}
The dataset will be distributed under a research license.

\paragraph{Have any third parties imposed IP-based or other restrictions on the data associated with the instances?}
No.

\paragraph{Do any export controls or other regulatory restrictions apply to the dataset or to individual instances?}
NA.

\subsection{Maintenance}

\paragraph{Who is supporting/hosting/maintaining the dataset?}
The authors will maintain the dataset. In particular, Laura Hanu (

\paragraph{How can the owner/curator/manager of the dataset be contacted (e.g., email address)?}
The website of the dataset will contain all information to contact the authors and or maintainers of the dataset.

\paragraph{Is there an erratum?}
No.

\paragraph{Will the dataset be updated (e.g., to correct labeling errors, add new instances, delete instances)?}
Yes, the website will contain a mechanism to version and update the dataset in case of errors.

\paragraph{If the dataset relates to people, are there applicable limits on the retention of the data associated with the instances (e.g., were individuals in question told that their data would be retained for a fixed period of time and then deleted)?}

\paragraph{Will older versions of the dataset continue to be supported/hosted/maintained?}
Yes through versioning on GitHub.

\paragraph{If others want to extend/augment/build on/contribute to the dataset, is there a mechanism for them to do so?}
Yes, on the website.

\paragraph{Will these contributions be validated/verified?} 
Yes, by the authors and maintainers of the dataset.

\section{Dataset Examples}
In figures \cref{tab:datasetexamples1}, \cref{tab:datasetexamples2}, and \cref{tab:datasetexamples3} we show random examples of the dataset with two comments (or less if a video only received one comment).

\begin{table}[h!]
  \centering
  \begin{tabular}{|c | p{3.5cm} | p{3.5cm} | p{3.5cm} |} \hline
    \textbf{Video} & \textbf{Title} & \textbf{Comment} & \textbf{Comment} \\ \hline

\raisebox{-0.5\totalheight}{\includegraphics[width=10mm]{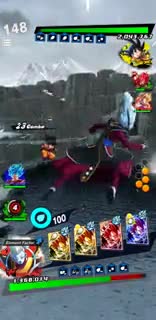}} & Beerus and Whis are still a deadly combo! (Zenkai 3) & So do I, by far one of my favorite units. Thanks! & Really well played, love seeing Beerus in action! \\ \hline
\raisebox{-0.5\totalheight}{\includegraphics[width=10mm]{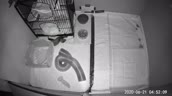}} & This little demon falls out of bed multiple times per night and just keeps snoozing. & It's pretty funny, I have dozens of these at this point ...! & - \\ \hline
\raisebox{-0.5\totalheight}{\includegraphics[width=10mm]{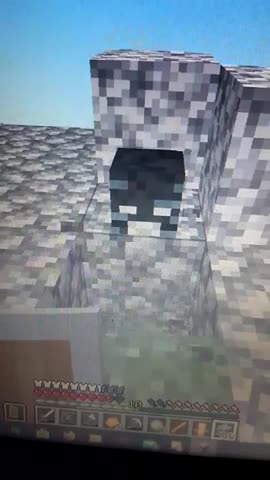}} & Anyone knows what this is? I was playing on an private Nitrado Minecraft Server with my 2 Friends, we were playing on the earliest version. When we were building, we realized that on a random mountain this skull just appeared, we couldn't destroy it and it just spawned every kind of mob & idk man, some kind of glitch maybe? & we were on peaceful \\ \hline
\raisebox{-0.5\totalheight}{\includegraphics[width=10mm]{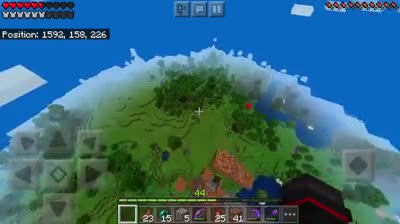}} & I know it’s not that crazy but I’m still really proud:) & Pressure and I didn’t want to fly all the way up again, I also was supposed & Why ender pearl? \\ \hline
\raisebox{-0.5\totalheight}{\includegraphics[width=10mm]{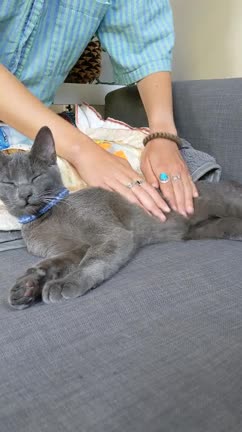}} & Gotta KNEAD the dough & That sound ahh so cute lol & - \\ \hline
\raisebox{-0.5\totalheight}{\includegraphics[width=10mm]{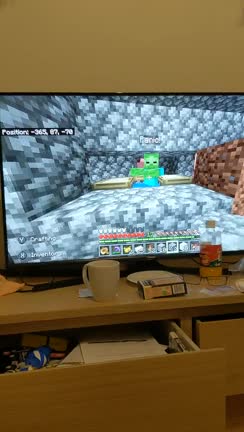}} & Anyone know why my iron farm won't work? They have beds & You built a java iron farm on bedrock lol & Do they have work stations? \\ \hline
\raisebox{-0.5\totalheight}{\includegraphics[width=10mm]{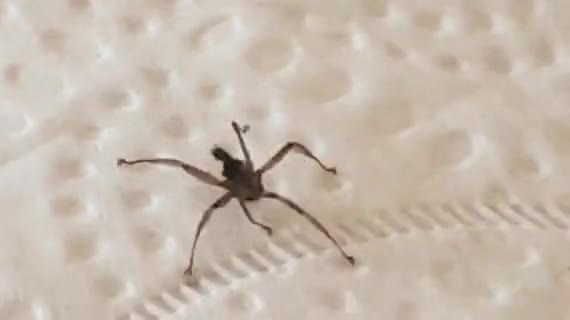}} & What am I looking at here? ',:/ & That looks like some sort of flightless fly. It kind of looks like a bat fl & - \\ \hline

  \end{tabular}
  \caption{A set of random samples from the dataset, showing title and up to two comments per video. (Included here since the guidelines only allow pdf/mp4 supplement)}\label{tab:datasetexamples1}
\end{table}

\begin{table}[h!]
  \centering
  \begin{tabular}{|c | p{3.5cm} | p{3.5cm} | p{3.5cm} |} \hline
    \textbf{Video} & \textbf{Title} & \textbf{Comment} & \textbf{Comment} \\ \hline
\raisebox{-0.5\totalheight}{\includegraphics[width=10mm]{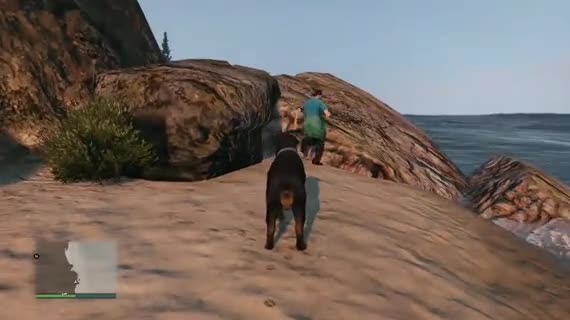}} & This happened to me while running around as a dog, and I feel like the music is fitting & what is the music i like that & - \\ \hline
\raisebox{-0.5\totalheight}{\includegraphics[width=10mm]{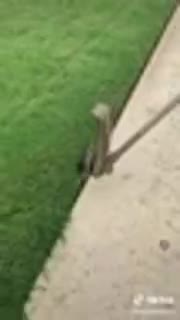}} & A cutting edge Tik Tok! & How does one obtain this power? & That’s really steady \\ \hline
\raisebox{-0.5\totalheight}{\includegraphics[width=10mm]{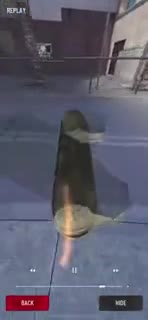}} & Didn’t notice the change in physics until trying this gap again & Yeah, finding a lot of spots with strange physics, but it's mostly because  & - \\ \hline
\raisebox{-0.5\totalheight}{\includegraphics[width=10mm]{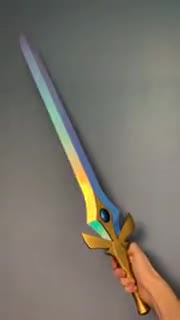}} & [self] I made my own She-Ra Sword using iridescent vinyl! & Its a PLA 3D print with aluminium core & What's underneath the vinyl? \\ \hline
\raisebox{-0.5\totalheight}{\includegraphics[width=10mm]{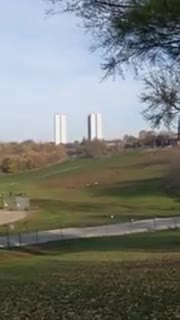}} & 3 minutes of hill sprints in 17 seconds at Don Valley Park East this morning. Can you feel the burn??? & Funny. I guess being corrected counts as “attitude” now. & Oh my...the contempt! Lmao....From your attitude I bet no one tells you any \\ \hline
\raisebox{-0.5\totalheight}{\includegraphics[width=10mm]{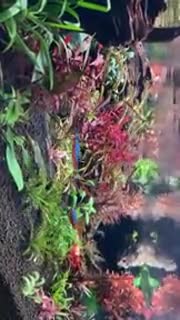}} & "Hidden Pools" 1 & That tnak is beautiful I love the plants and colours! & - \\ \hline
\raisebox{-0.5\totalheight}{\includegraphics[width=10mm]{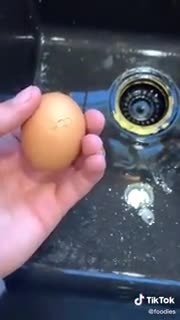}} & Bad egg from Walmart & Rough crowd. & Did you read his comment until the end? \\ \hline
\raisebox{-0.5\totalheight}{\includegraphics[width=10mm]{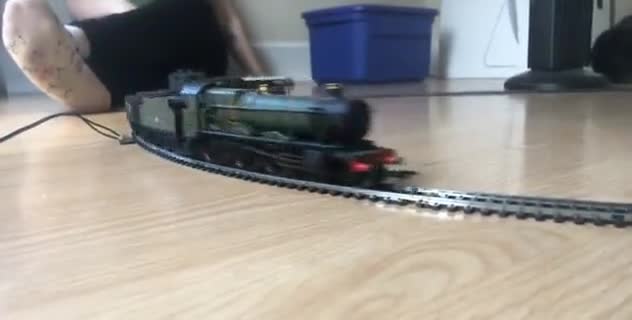}} & First trip around the track & Those were my guesses as it seems to be on the smaller side of the 4-6-0 cl & A Manor or a Hall? \\ \hline
\raisebox{-0.5\totalheight}{\includegraphics[width=10mm]{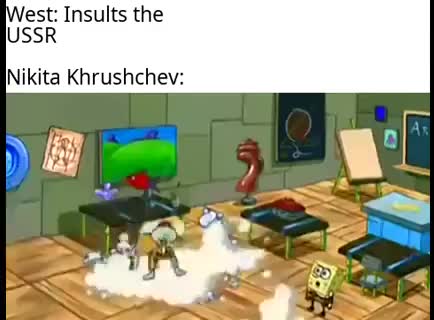}} & Ra Ra Rasputin Russia's greatest rage machine & Hello everyone! We have opened new  & lrrelevant title but the meme is ok \\ \hline
\raisebox{-0.5\totalheight}{\includegraphics[width=10mm]{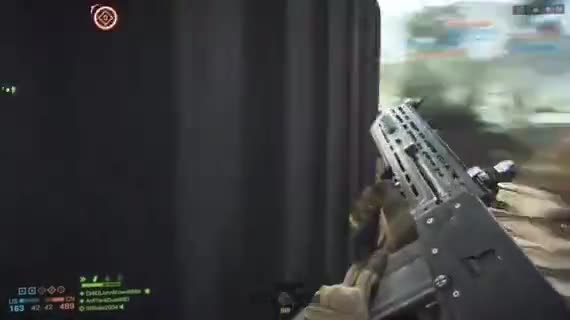}} & Anyone else hate having to ADS? & My friends and I love Marty Robbins XD thank you for this & "And the stranger's aim was deadly with the big iron on his hip." \\ \hline

\end{tabular}
  \caption{A set of random samples from the dataset, showing title and up to two comments per video. (Included here since the guidelines only allow pdf/mp4 supplement)}\label{tab:datasetexamples2}
\end{table}

\begin{table}[h!]
  \centering
  \begin{tabular}{|c | p{3.5cm} | p{3.5cm} | p{3.5cm} |} \hline
    \textbf{Video} & \textbf{Title} & \textbf{Comment} & \textbf{Comment} \\ \hline
\raisebox{-0.5\totalheight}{\includegraphics[width=10mm]{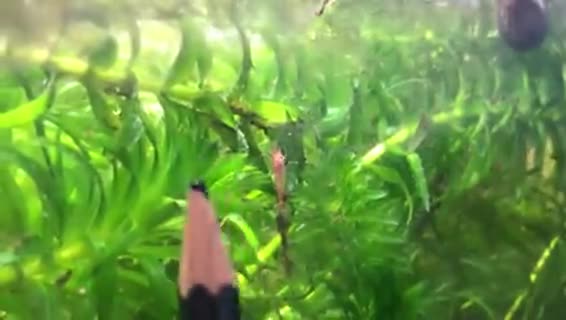}} & Have had 6 cherry shrimps for 2 months, always can’t find them and thought they are dead / been eaten. Surprised to see this baby shrimp today! Pencil for scale in video. & That’s how they hide from me! Thanks for your advice. & They are very good at hiding. If there are holes or pits in your substrate  \\ \hline
\raisebox{-0.5\totalheight}{\includegraphics[width=10mm]{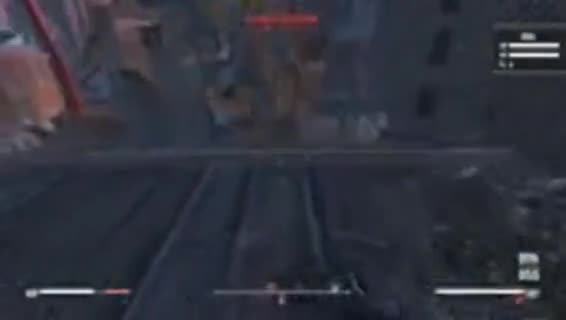}} & The preferable option & Ugh. & Aaaiiee! \\ \hline
\raisebox{-0.5\totalheight}{\includegraphics[width=10mm]{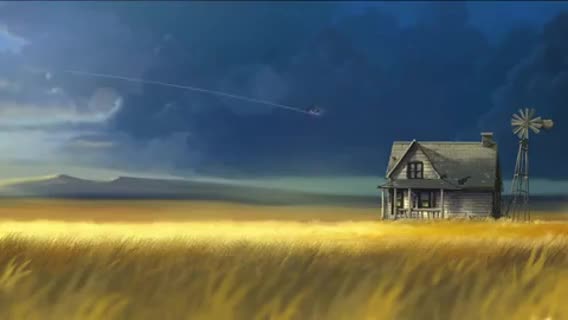}} & Is he strong? Listen, Bud! He's got radioactive blood. Can he swing from a thread? Take a look overhead. & Is that the house from Courage the Cowardly Dog & Oh, wheat! \\ \hline
\raisebox{-0.5\totalheight}{\includegraphics[width=10mm]{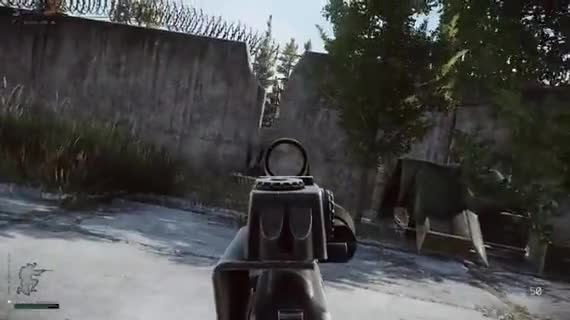}} & Pog gloryhole shot sorry I got excited I killed him and I finished my punisher pt3 in that raid as well & KomodoHype & - \\ \hline
\raisebox{-0.5\totalheight}{\includegraphics[width=10mm]{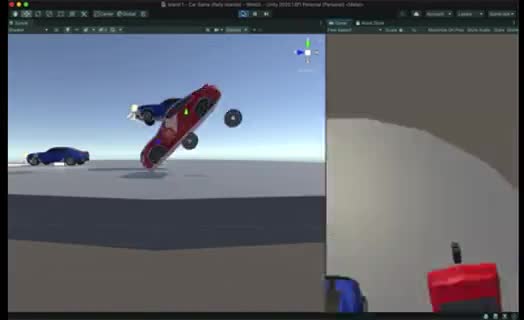}} & Why does this keep happening? (I am a noob) & On the rigidbody attached to your object select freeze rotation of the axis & When I was trying to fix it yesterday I found out that I didn’t have a mesh \\ \hline
\raisebox{-0.5\totalheight}{\includegraphics[width=10mm]{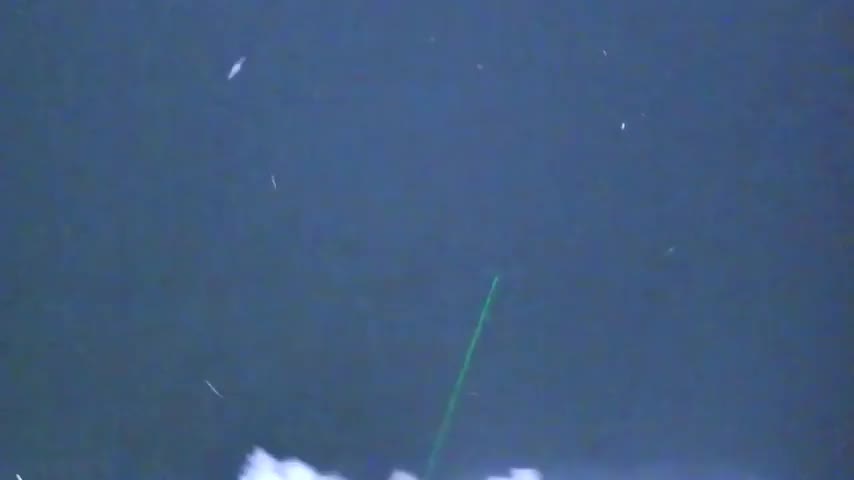}} & I'm sure you've all seen this one, but just incase you haven't & A bug that flinches when hit by laser. & - \\ \hline
\end{tabular}
  \caption{A set of random samples from the dataset, showing title and up to two comments per video. (Included here since the guidelines only allow pdf/mp4 supplement)}\label{tab:datasetexamples3}
\end{table}

\clearpage



\end{document}